\pdfoutput=1
\documentclass[10pt,twocolumn,letterpaper]{article}
\PassOptionsToPackage{table}{xcolor}
\usepackage[OT1]{fontenc}
\usepackage{newunicodechar}
\newunicodechar{–}{--}
\usepackage[pagenumbers]{cvpr}

\def\confName{CVPR}
\def\confYear{2027}

\usepackage{amsmath,amsfonts,bm}

\def\eqref#1{equation~\ref{#1}}
\def\1{\bm{1}}

\DeclareMathAlphabet{\mathsfit}{\encodingdefault}{\sfdefault}{m}{sl}
\SetMathAlphabet{\mathsfit}{bold}{\encodingdefault}{\sfdefault}{bx}{n}

\usepackage{graphicx}
\usepackage{booktabs}
\usepackage{array}
\usepackage{multirow}
\usepackage{subcaption}
\usepackage{placeins}
\usepackage{xcolor}
\usepackage{amssymb}

\definecolor{cvprblue}{RGB}{22, 68, 120}
\definecolor{humiblue}{RGB}{22, 68, 120}
\definecolor{humisub}{RGB}{52, 66, 84}
\definecolor{humirule}{RGB}{192, 204, 218}
\definecolor{humimuted}{RGB}{118, 128, 140}
\definecolor{LightGray}{gray}{0.93}

\usepackage[pagebackref,breaklinks,colorlinks,allcolors=humiblue]{hyperref}
\usepackage{url}

\makeatletter
\newif\ifhumi@appendix
\humi@appendixfalse
\newif\ifhumi@supptitlepage
\humi@supptitlepagefalse

\def\maketitlesupplementary{%
   \newpage
   \global\humi@supptitlepagetrue
   \twocolumn[{%
     \centering
     {\Large\bf \thetitle\par}\vspace{0.45em}%
     {\large\bf Supplementary Material\par}\vspace{5pt}%
     {\color{humirule}\rule{0.92\textwidth}{0.5pt}}\par\vspace{0.8em}%
   }]%
}

\AddToShipoutPicture{%
  \AtTextUpperLeft{%
    \ifnum\value{page}=1\relax
    \else
      \ifhumi@supptitlepage
        \global\humi@supptitlepagefalse
      \else
        \put(0,\LenToUnit{0.26in}){%
          \parbox{\textwidth}{%
            \footnotesize\color{humimuted}%
            \textit{Beyond Decoupled Control: Rethinking Whole-Body Navigation for Humanoid Robots}%
            \hfill
            \ifhumi@appendix
              \textup{HumiAct -- Supplementary Material}%
            \else
              \textup{HumiAct}%
            \fi
            \par\vspace{-2pt}%
            {\color{humirule}\rule{\textwidth}{0.4pt}}%
          }%
        }%
      \fi
    \fi
  }%
}
\makeatother

\newcommand{\affmark}[1]{\textsuperscript{\textcolor{humiblue}{#1}}}
\newcommand{\unilogobox}[1]{\hspace{0.12em}\raisebox{-0.30em}[0pt][0pt]{#1}\hspace{0.24em}}
\newcommand{\honorlogobox}[1]{\hspace{0.14em}\raisebox{-0.18em}[0pt][0pt]{#1}\hspace{0.24em}}

\newcommand{\mainref}[1]{\ref{#1}}

\newcommand{\grt}{GR00T$^{\dagger}$}

\title{Beyond Decoupled Control: Rethinking Whole-Body Navigation for Humanoid Robots}

\author{%
\raisebox{18pt}[\dimexpr\height-6pt\relax][\dimexpr\depth-30pt\relax]{%
\parbox[t]{0.96\textwidth}{\centering
\href{https://fayo.ai}{\textcolor{black}{Haihong Hao}}\affmark{1,2} \quad Hui Zhu\affmark{2} \quad Tingda Zhuang\affmark{2} \quad Wenwen Luo\affmark{3,2} \quad Chen Liang\affmark{4,2} \\[1pt]
Lei Chen\affmark{1} \quad Mingfei Han\affmark{1} \quad Xiaojun Chang\affmark{1} \quad Zhihui Li\affmark{1} \\[3.5pt]
{\normalsize\color{humisub}%
\affmark{1}\unilogobox{\includegraphics[height=1.26em]{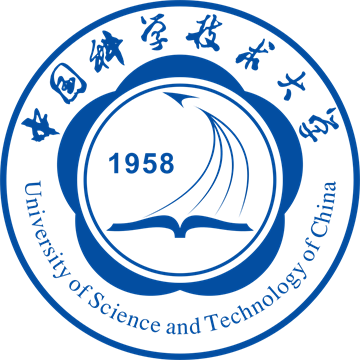}}University of Science and Technology of China \qquad \affmark{2}\honorlogobox{\includegraphics[height=1.02em]{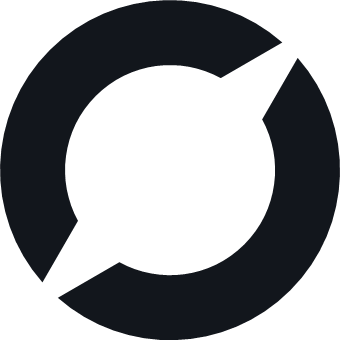}}Honor \\[2.5pt]
\affmark{3}\unilogobox{\includegraphics[height=1.26em]{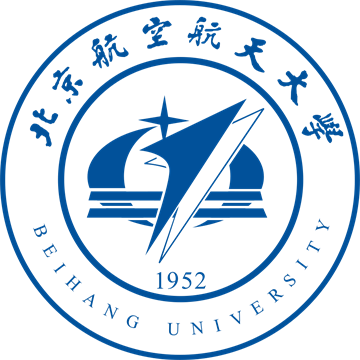}}Beihang University \qquad \affmark{4}\unilogobox{\includegraphics[height=1.26em]{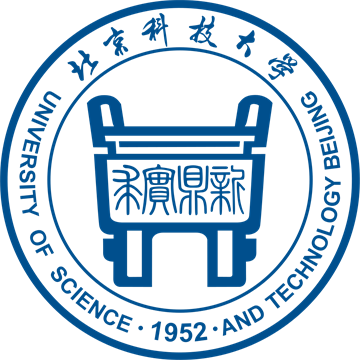}}University of Science and Technology Beijing}}}%
}

\begin{document}

\maketitle

\begin{abstract}
Humanoid robots hold promise for general-purpose embodied intelligence, yet their navigation systems largely inherit decoupled paradigms from wheeled and quadrupedal platforms.
While such modular designs enable scalable optimization, they underexploit a key advantage of humanoids: the tight coupling of perception, locomotion, and physical interaction.
Realizing this potential requires rethinking the conventional separation between navigation and manipulation.
In this work, we revisit whole-body humanoid navigation and present \emph{HumiAct}, a unified framework that couples a high-level reasoning planner with a shared vision-language-action policy and whole-body controller for coordinated planning, locomotion, and manipulation.
We establish a training and evaluation ecosystem spanning heterogeneous datasets and physics-based simulation with paired-route protocols.
Paired runs on the same routes reveal a \emph{vision dependence dilemma}.
Camera feedback helps the humanoid stay on course, yet changing image appearance alone degrades performance due to reliance on scene-specific cues.
We introduce \emph{latent action guidance} to distill action-conditioned motion priors from cross-embodiment navigation and manipulation trajectories into the shared policy, reducing appearance sensitivity without inference overhead.
Built on GR00T, HumiAct's single shared policy outperforms fine-tuned baselines and separately trained specialists on direct navigation, planner-guided navigation, manipulation, and tasks that combine navigation with physical interaction.
We further validate the system on a real humanoid across navigation and interaction tasks.
Together, HumiAct and its data-to-robot ecosystem advance humanoid navigation toward unified whole-body intelligence.
\end{abstract}
\vspace{-10pt}
\section{Introduction}
\label{sec:introduction}

Humanoid robots combine whole-body mobility with the ability to physically interact with their surroundings, a capability brought increasingly within reach by recent advances in humanoid control and generalist robot policies~\cite{fu2024humanplus,nvidia2025grootn1}.
Yet many humanoid systems still treat navigation and manipulation as separate problems.
While this separation simplifies individual skill development, it underexploits a defining advantage of humanoid embodiment: coordinating perception, locomotion, and physical interaction through the same body.
A humanoid approaching an object must coordinate where it moves, what it sees, and how it configures its body for the interaction that follows.
These demands motivate a broader question:
\emph{How can a humanoid perceive, move, and physically interact with its environment as a unified embodied agent?}

\begin{figure*}[t]
    \centering
    \includegraphics[width=0.89\textwidth]{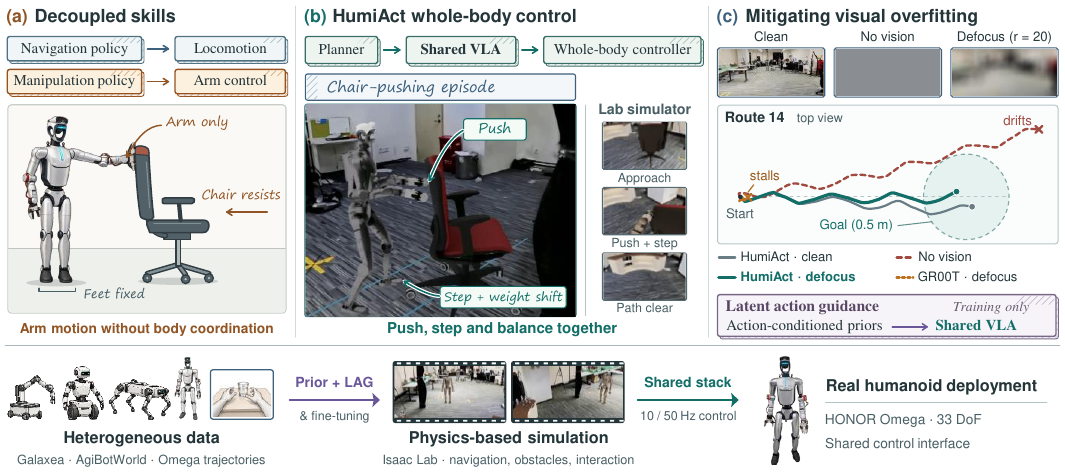}
    \vspace{-2mm}
    \caption{\textbf{Motivation and overview of HumiAct.}
\textbf{(a) Decoupled skills:} separate navigation and manipulation stacks limit whole-body coordination, illustrated by arm-only chair pushing with fixed feet.
\textbf{(b) Whole-body control:} HumiAct connects a task planner, a shared VLA policy, and a whole-body controller to coordinate pushing, stepping, and balance (chair-pushing episode).
\textbf{(c) Mitigating visual overfitting:} Under strong defocus, HumiAct follows nearly the same path and reaches nearly the same endpoint as with clean images.}
    \label{fig:intro_overview}
    \vspace{-3mm}
\end{figure*}

Existing deployments typically pair separate navigation and manipulation policies with partial-body controllers~\cite{liu2024okrobot}, imposing boundaries between behaviors executed by the same body (Fig.~\ref{fig:intro_overview}a).
We present \emph{HumiAct}, a three-level humanoid agent that unifies navigation and manipulation within a shared policy--controller stack under high-level task planning (Fig.~\ref{fig:intro_overview}b).
A language-model planner interprets instructions, decomposes long-horizon tasks into short language subtasks, and monitors progress.
Rather than switching between separate skill modules across subtasks, one shared vision-language-action (VLA) policy maps the active subtask, visual observations, and proprioception to whole-body motion commands, while one low-level whole-body controller tracks these commands and maintains balance.
This design eliminates policy--controller handoffs between navigation and manipulation and supports coupled loco-manipulation such as stepping while pushing.

Unifying navigation and manipulation in a mobile VLA policy, however, exposes a \emph{vision dependence dilemma}.
While whole-body traversal and interaction span wide viewpoint and background shifts, humanoid demonstrations remain scarce and confined to few scenes.
Removing vision causes accumulated drift and trajectory deviation, whereas incorporating vision enables closed-loop correction relative to the environment.
Yet when trained on limited humanoid scenes, the policy readily latches onto static appearance shortcuts correlated with training actions, degrading sharply under unfamiliar scenes or image corruptions.
The central challenge is to preserve closed-loop visual feedback without overfitting to scene-specific appearance.
We address this challenge by grounding visual representations in \emph{action-centric predictive priors} learned from diverse cross-embodiment trajectories.
Predicting how actions change visual observations pushes features to capture task-relevant geometry and ego-motion rather than static background textures, while cross-embodiment navigation and manipulation datasets provide the scene diversity missing from humanoid-only data.
Although different embodiments use distinct control interfaces, their motions can be abstracted into shared patterns of how actions transform camera views.
We therefore introduce \emph{latent action guidance} (LAG) to transfer these motion priors into the shared VLA policy.
Mapping heterogeneous robot trajectories onto a compact vocabulary of shared latent actions, an action-conditioned predictor first learns across embodiments how each behavior alters what the camera sees; its latent current and predicted future features are then distilled into the VLA backbone prior to whole-body fine-tuning, shaping action-aware visual representations offline without adding any inference-time module (Fig.~\ref{fig:intro_overview}c).
Evaluated in a physics-based reconstruction of an indoor workspace with paired-route protocols, HumiAct outperforms fine-tuned GR00T by 20 percentage points in success rate on obstacle routes and by 9.4 points on combined navigation--interaction tasks, while averaging 84\% success across five image corruptions on goal-facing routes.
We further deploy HumiAct on a physical humanoid robot, demonstrating unified navigation and whole-body interaction in the real world.

Our contributions are threefold:
\begin{itemize}
    \item \textbf{A unified humanoid agent for navigation and interaction.}
    HumiAct couples a language-model planner with one shared VLA policy and one whole-body controller, so that navigation, manipulation, and the transition between them are produced by the same embodied agent.

    \item \textbf{Action-centric guidance for visual control.}
    We identify the vision dependence dilemma and introduce LAG, which transfers motion priors learned from heterogeneous embodied trajectories into the shared VLA policy to reduce reliance on scene-specific appearance.

    \item \textbf{A data-to-robot ecosystem.}
    We build a training and evaluation pipeline spanning heterogeneous embodied data, a physics-based reconstruction of an indoor workspace, and real-humanoid deployment, with paired-route protocols that isolate policy, planner, and visual-input effects.
\end{itemize}

\begin{figure*}[t]
    \centering
    \includegraphics[width=0.84\textwidth]{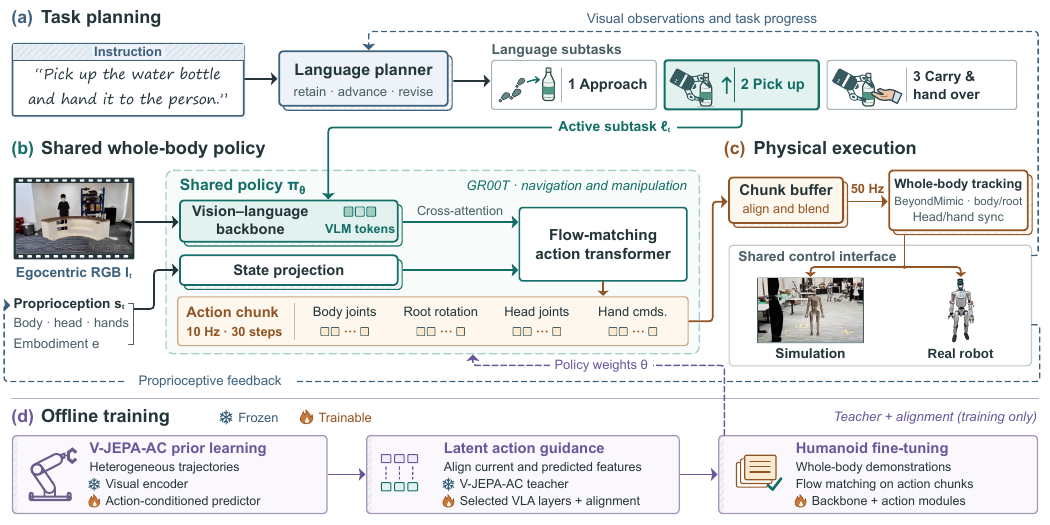}
    \vspace{-2mm}
    \caption{\textbf{Overview of HumiAct.}
    (a) A task planner updates the active language subtask using visual progress.
    (b) On hardware, the shared VLA predicts 30-step whole-body action chunks at 10~Hz.
    (c) Buffered and blended references are streamed at 50~Hz for whole-body tracking and synchronized head and hand execution.
    (d) Predictive prior learning, LAG transfer, and humanoid fine-tuning provide the policy weights; the teacher and alignment modules are used only during training.}
    \label{fig:cha_pipeline}
    \vspace{-3mm}
\end{figure*}
\section{Related Work}
\label{sec:related}
\textbf{Navigation and mobile manipulation.}
Vision-and-language navigation has progressed from graphs to continuous spaces~\citep{anderson2018vln,krantz2020vlnce} via large-scale simulation and RL~\citep{savva2019habitat,wijmans2020ddppo}, advancing alignment, history reasoning, and multilingual grounding~\citep{fried2018speaker,wang2019rcm,hong2021vlnbert,chen2021hamt,ku2020rxr,hao2026conav,han2026implicit}, alongside cross-embodiment goal-conditioned navigation~\citep{shah2023gnm,shah2023vint,sridhar2023nomad}.
Language planners ground instructions into skills via affordances, feedback, code, or value maps~\citep{ahn2022saycan,driess2023palme,huang2022inner,liang2023codepolicies,huang2023voxposer}, while mobile manipulators pair navigation with grasping~\citep{liu2024okrobot,fu2024mobilealoha}.
Unlike wheeled or modular stacks, HumiAct unifies navigation and manipulation in one visuomotor policy and whole-body controller.\\
\textbf{Whole-body humanoid control and VLA policies.}
Motion priors and legged controllers~\citep{peng2018deepmimic,peng2021amp,peng2022ase,hwangbo2019agile,rudin2022walk,miki2022perceptive} underpin humanoid sim-to-real locomotion~\citep{gu2024humanoidgym}, teleoperation and imitation~\citep{he2024h2o,he2024omnih2o,fu2024humanplus}, and motion tracking~\citep{he2025hover,liao2025beyondmimic,luo2026sonic}.
Meanwhile, generalist VLA policies~\citep{brohan2023rt1,zitkovich2023rt2,ghosh2024octo,black2024pi0,kim2024openvla,nvidia2025grootn1} leverage large cross-robot datasets~\citep{oneill2024rtx,dasari2019robonet,walke2023bridge,khazatsky2024droid}.
Recent humanoid systems explore video-learned loco-manipulation~\citep{jiang2025wholebodyvla}, cluttered whole-body navigation~\citep{li2026tango}, and embodiment-routed stereo~\citep{wu2026eatrstereo}; HumiAct instead unifies planner-guided navigation, manipulation, and their composition in one policy while explicitly regulating its visual dependence.\\
\textbf{Latent actions, predictive learning, and visual generalization.}
Prior methods compress multimodal behavior or infer visual latent actions~\citep{lynch2020latentplans,shafiullah2022bet,lee2024vqbet,wang2023mimicplay,bruce2024genie,ye2025lapa}, learn latent world models~\citep{ha2018worldmodels,hafner2019planet,hafner2020dreamer,hafner2021dreamerv2,hao2026seer}, or guide policies with predictive video, state, denoising, and view-invariant representations~\citep{bardes2024vjepa,assran2025vjepa2,tian2024seer,zheng2026motionwam,jeong2026vila,hao2026latentpilot}.
Visual overfitting is typically countered via randomization, augmentation, or pretrained encoders~\citep{zhang2018overfitting,cobbe2019generalization,tobin2017domainrandomization,laskin2020rad,yarats2021drq,nair2022r3m,majumdar2024vc1} alongside self-, language-, and masked-supervised pretraining~\citep{caron2021dino,oquab2024dinov2,radford2021clip,he2022mae,assran2023ijepa,radosavovic2023mvp}.
LAG is complementary: conditioning prediction on five shared cross-embodiment behaviors avoids equating raw control vectors while distilling current and future features into the VLA without pixel prediction or test-time modules.

\section{Method}
\label{sec:method}
\subsection{Unified Whole-Body Control}
\label{sec:unified_control}
For a humanoid, approaching a target couples locomotion, viewpoint, body orientation, and the posture required for the interaction that follows.
HumiAct captures this coupling with the three levels of Fig.~\ref{fig:cha_pipeline}; navigation and manipulation share both the policy and the execution stack, so their transition is part of one continuous perception--action process.
\\
\textbf{Task planning with progress feedback.}
Given a long-horizon instruction $g$ and an initial observation $I_{t_0}$, a language-model planner produces an ordered sequence of short language subtasks,
\begin{equation}
    \mathcal{P}_0=(\ell_1,\ldots,\ell_M)=\Pi_{\mathrm{plan}}(g,I_{t_0}),
    \label{eq:task_plan}
\end{equation}
such as moving forward, turning, or interacting with an object.
It periodically evaluates progress from incoming observations and the execution context, and retains the current subtask, advances to the next one, or revises the remaining plan.
The subtask $\ell_t$ active at time $t$ is the language condition of the shared policy.
\begin{figure*}[t]
    \centering
    \includegraphics[width=0.84\textwidth]{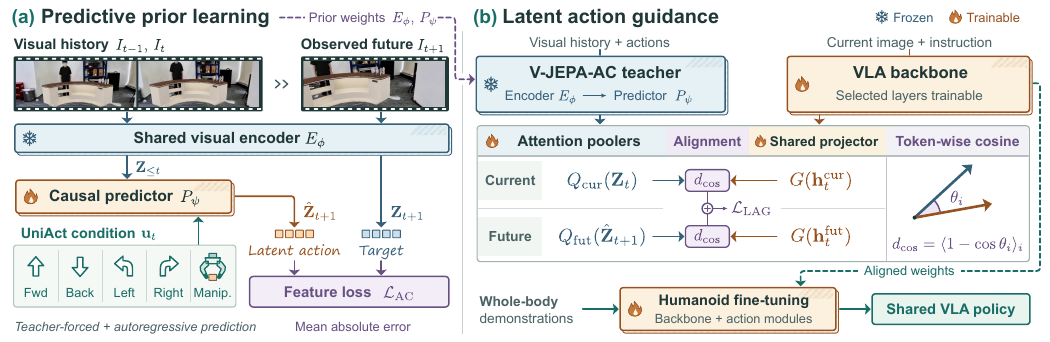}
    \vspace{-2mm}
       \caption{\textbf{Learning and transferring action-conditioned motion priors.}
    \textbf{(a) Predictive prior learning:} heterogeneous transitions mapped to five shared behaviors supervise causal predictor $P_\psi$ over frozen encoder $E_\phi$ via teacher-forced and autoregressive rollouts.
    \textbf{(b) Latent action guidance:} frozen V-JEPA-AC features pooled by $Q_{\mathrm{cur}}$ and $Q_{\mathrm{fut}}$ supervise projected VLA auxiliary representations ($G$) via cosine alignment.
    \textbf{Bottom:} the aligned backbone and action modules are fine-tuned on humanoid demonstrations; teacher, auxiliary tokens, and alignment modules are discarded at inference.}
    \label{fig:uniact_lag}
    \vspace{-5mm}
\end{figure*}
\\
\noindent\textbf{Shared whole-body action generation.}
Let $I_t$ be the egocentric RGB observation, $s_t$ the proprioceptive state of body, head, and hands, and $e$ the embodiment identity.
Following action-sequence policy designs~\citep{zhao2023act,chi2023diffusion}, one policy predicts an action chunk of horizon $H$ for every behavior:
\begin{equation}
    \hat{\mathbf A}_t=\hat{\mathbf a}_{t:t+H-1}=\pi_\theta(c_t),
    \qquad
    c_t=(I_t,s_t,\ell_t,e).
    \label{eq:shared_whole_body_policy}
\end{equation}
We build on the GR00T architecture~\citep{nvidia2025grootn1}: a vision-language backbone, embodiment-conditioned state and action projections, and an action transformer $f_\theta$ trained with flow matching~\citep{lipman2023flow}.
For a demonstration chunk $\mathbf A$, noise $\boldsymbol\epsilon\sim\mathcal N(0,\mathbf I)$, and $\tau\in[0,1]$, the noised action and its target velocity are
\begin{equation}
    \mathbf X_\tau=(1-\tau)\boldsymbol\epsilon+\tau\mathbf A,
    \qquad
    \mathbf V^\star=\mathbf A-\boldsymbol\epsilon,
    \label{eq:action_flow_path}
\end{equation}
and the policy is trained with
\begin{equation}
    \mathcal{L}_{\mathrm{FM}}
    =
    \mathbb{E}
    \left[
        \frac{\left\|\mathbf m\odot\left(f_\theta(\mathbf X_\tau,\tau,c_t)-\mathbf V^\star\right)\right\|_F^2}{\|\mathbf m\|_1}
    \right],
    \label{eq:flow_matching}
\end{equation}
where the mask $\mathbf m$ selects valid action dimensions and steps; at inference, integrating $f_\theta$ from $\mathbf X_0=\boldsymbol\epsilon$ over $\tau\in[0,1]$ yields $\hat{\mathbf A}_t$.
Because legs, heading, head, waist, and arms are predicted within one trajectory, an interaction subtask continues from the body configuration and viewpoint that the approach produced.
\\
\textbf{Whole-body references and multi-rate execution.}
Each predicted step is decoded into a synchronized reference
\begin{equation}
    \mathbf r_{t+j}=\left(\hat{\mathbf q}^{\mathrm{body}}_{t+j},\hat{\mathbf R}^{\mathrm{root}}_{t+j},\hat{\mathbf q}^{\mathrm{head}}_{t+j},\hat{\mathbf g}_{t+j}\right),
    \quad j=0,\ldots,H-1,
    \label{eq:whole_body_reference}
\end{equation}
of body-joint positions, root orientation, head-joint positions, and hand commands.
On hardware, inference and execution are asynchronous: chunks predicted at 10~Hz are buffered, aligned to the execution timeline, and blended at transitions, and references are streamed at 50~Hz to a whole-body controller that tracks body and root targets from proprioceptive feedback and maintains balance, with head and hand commands executed in sync.

\subsection{Learning Latent Actions from Heterogeneous Trajectories}
\label{sec:latent_action_learning}

Embodiments differ in their control interfaces, yet their trajectories share one learning signal: how actions shape visual dynamics (Fig.~\ref{fig:uniact_lag}a).
\textbf{Unified action semantics (UniAct).}
From the motion records and task annotations of open robot datasets~\citep{jiang2025galaxea,bu2025agibot} and our humanoid trajectories, we map every transition to one of five behaviors: moving forward, moving backward, turning left, turning right, and manipulation.
Base translation and rotation identify the navigation classes, arm and hand motion identify manipulation, and idle or ambiguous intervals are discarded.
The resulting one-hot condition $\mathbf u_t$ gives a visual transition the same action meaning on every embodiment.
It is used only as the condition for predictive learning; physical execution keeps the continuous whole-body commands of Sec.~\ref{sec:unified_control}.

\paragraph{Action-conditioned future prediction.}
We build V-JEPA-AC on the predictive architecture of V-JEPA~2~\citep{assran2025vjepa2}.
A frozen pretrained encoder $E_\phi$ defines the representation space and a causal predictor $P_\psi$ models its evolution under the semantic condition:
\begin{equation}
    \mathbf Z_t = E_\phi(I_t),
    \qquad
    \hat{\mathbf Z}_{t+1}=P_\psi\left(\mathbf Z_{\leq t},\mathbf u_{\leq t}\right),
    \label{eq:action_conditioned_prediction}
\end{equation}
with normalized patch-token representations $\mathbf Z_t$.
Predictions are supervised by encoded future observations,
\begin{equation}
    \mathcal{L}_{\mathrm{AC}}
    =
    \mathbb{E}\left[
        d\!\left(\hat{\mathbf Z}^{\mathrm{TF}}_{1:T-1},\mathbf Z_{1:T-1}\right)
        +
        d\!\left(\hat{\mathbf Z}^{\mathrm{AR}}_{1:K},\mathbf Z_{1:K}\right)
    \right],
    \label{eq:action_conditioned_loss}
\end{equation}
where $d$ is the mean absolute feature difference and the two terms cover teacher-forced prediction and $K$-step autoregressive rollouts.
The predictor's final-layer tokens, projected into the encoder feature space, are the latent actions: they encode what the scene will look like if the given action is executed.

\subsection{Latent Action Guidance}
\label{sec:latent_action_guidance}

LAG transfers these action-conditioned dynamics into the shared VLA (Fig.~\ref{fig:uniact_lag}b).
Following knowledge and intermediate-feature distillation~\citep{hinton2015distill,romero2015fitnets,zagoruyko2017attention}, the frozen teacher provides current features $\mathbf Z_t$ and predicted future features $\hat{\mathbf Z}_{t+1}$.
Given the current image and instruction, the VLA produces auxiliary-token representations $\mathbf h_t^{\mathrm{cur}}$ and $\mathbf h_t^{\mathrm{fut}}$ at an intermediate backbone layer.
A shared projector $G$ maps them into the teacher space, learned attention poolers $Q_{\mathrm{cur}}$ and $Q_{\mathrm{fut}}$ form compact teacher targets, and we minimize
\begin{equation}
    \begin{aligned}
        \mathcal{L}_{\mathrm{LAG}}
        = \mathbb{E}\bigl[
        &d_{\cos}\!\left(G(\mathbf h_t^{\mathrm{cur}}),Q_{\mathrm{cur}}(\mathbf Z_t)\right) \\
        {}+{}&
        d_{\cos}\!\left(G(\mathbf h_t^{\mathrm{fut}}),Q_{\mathrm{fut}}(\hat{\mathbf Z}_{t+1})\right)
        \bigr],
    \end{aligned}
    \label{eq:latent_action_guidance}
\end{equation}
with $d_{\cos}$ the mean token-wise cosine distance.
The backbone thus learns to represent the current scene together with its expected evolution under the instructed behavior, a target that is defined by motion and not by the appearance of a particular environment.
The aligned backbone then initializes the shared VLA, which is trained for whole-body control with $\mathcal{L}_{\mathrm{FM}}$.
Teacher, auxiliary tokens, and alignment modules are discarded afterwards, so LAG adds no inference cost and leaves the policy interface of Eq.~\ref{eq:shared_whole_body_policy} unchanged.

\subsection{Data-to-Robot Ecosystem}
\label{sec:data_to_robot}

Our ecosystem connects heterogeneous motion learning, humanoid adaptation, and closed-loop evaluation through a common observation--action interface.\\
\begin{table}[t]
    \centering
    \footnotesize
    \setlength{\tabcolsep}{3.5pt}
    \caption{\textbf{Training data.} Stage-specific episode counts. All sources cover navigation and manipulation; the whole-body demonstrations additionally cover navigation--interaction.}
    \label{tab:training_data}
    \vspace{-2mm}
    \begin{tabular*}{\linewidth}{@{\extracolsep{\fill}}llrl@{}}
        \toprule
        Source & Embodiment & Episodes & Stage \\
        \midrule
        Galaxea~\citep{jiang2025galaxea} & R1 Lite / Pro & 34,353 & Prior, LAG \\
        AgiBotWorld~\citep{bu2025agibot} & A2D & 142,069 & Prior, LAG \\
        Humanoid trajectories & Target humanoid & 1,962 & Prior, LAG \\
        Whole-body demos & Target humanoid & 5,421 & Fine-tuning \\
        \bottomrule
    \end{tabular*}
    \vspace{-3mm}
\end{table}
\noindent\textbf{Multi-source data.}
We combine navigation and manipulation trajectories from open robot datasets with whole-body humanoid demonstrations (Table~\ref{tab:training_data}).
The heterogeneous trajectories, about 178k episodes from three embodiments, provide broad coverage of action-dependent visual dynamics for prior learning and transfer.
The humanoid demonstrations then connect these priors to coordinated locomotion, object handling, and navigation--interaction on the target platform. \textbf{Training pipeline.}
Training has three stages.
(i)~The predictor $P_\psi$ is trained with $\mathcal{L}_{\mathrm{AC}}$ on the heterogeneous corpus, with the pretrained V-JEPA~2 encoder frozen.
(ii)~The frozen teacher transfers motion priors into a GR00T-initialized VLA through $\mathcal{L}_{\mathrm{LAG}}$ on the same corpus, updating selected backbone layers and the alignment modules.
(iii)~Backbone and action modules are jointly fine-tuned on the whole-body demonstrations with $\mathcal{L}_{\mathrm{FM}}$. All stages train in BF16 on 16 NVIDIA H20 GPUs: 5k prior steps (24-layer $P_\psi$, $K{=}2$), 32-token LAG alignment, and 200k fine-tuning steps (batch 96) under video-disjoint splits (Appendix).
\textbf{Physics-based simulation and task evaluation.}
We build the simulation environment on Isaac Lab~\citep{mittal2025isaaclab} with PhysX, combining a reconstructed indoor workspace with collision-enabled obstacles and manipulable objects.
We vary robot initialization, object placement, illumination, and object mass to evaluate behavior across visual and physical conditions, and seeded resets with automatic task checks give repeatable measurements of navigation progress, collisions, grasping, and object transport.
An adapter connects the unified motion references to a BeyondMimic tracker~\citep{liao2025beyondmimic} with synchronized head and hand execution; physics runs at 200~Hz and whole-body tracking at 50~Hz.
The simulator exposes the policy interface used on hardware, enabling closed-loop evaluation and paired re-runs of the same route under different policies or visual inputs.
\textbf{Real-humanoid deployment.}
We deploy HumiAct on the same 33-DoF humanoid platform used in EATR-Stereo~\citep{wu2026eatrstereo}. The robot has synchronized head-mounted cameras and a 37-dimensional proprioceptive state comprising joint positions and base orientation.
The shared VLA runs on an NVIDIA RTX 5000 Ada at 10~Hz and predicts 30-step action chunks; BF16 inference with four flow-integration steps keeps action generation within the control budget.
A ROS~2 client exchanges visual and proprioceptive observations with the planner--policy service through ZeroMQ and streams whole-body references at 50~Hz, while asynchronous buffering and overlap blending maintain continuous execution.

\section{Experiments}
\label{sec:experiments}

\subsection{Experimental Setup}
\label{sec:experimental_setup}
\textbf{Tasks and metrics.}
We evaluate the humanoid in the reconstructed indoor workspace (Sec.~\ref{sec:data_to_robot}).
\emph{Direct navigation} uses obstacle-free routes with a goal-facing start, a forward-motion instruction, a 10\,s episode, and a 0.5\,m goal radius.
\emph{Planner navigation} uses obstacle routes on which the planner selects and updates the instruction.
\emph{Manipulation} starts at the work surface, and \emph{navigation--interaction} uses farther starts that require an approach followed by the object task.
We report success rate (SR), success weighted by path length (SPL), and the fraction of trials with an obstacle contact above 10\,N (Coll.).
In simulation every policy executes the first 30 steps of each 40-step chunk at 20\,Hz. GPT-5.6 Terra is the default planner.\\
\textbf{Compared policies.}
All policies are fine-tuned on our humanoid demonstrations and share the whole-body controller.
\emph{GR00T}~\citep{nvidia2025grootn1} is fine-tuned from the public base model and represents a vision-only whole-body VLA.
GR00T uses the same 200k-step fine-tuning (batch 96). The visual-input study also evaluates \emph{\grt}, a second GR00T checkpoint fine-tuned with HumiAct's downstream recipe from the public base model.
The \emph{navigation-only} specialist shares HumiAct's LAG initialization but sees no manipulation data, the \emph{manipulation-only} specialist sees no navigation data, and the \emph{prior-initialized reference} is an earlier LAG-initialized policy without the final joint adaptation.

\begin{table*}[t]
\centering
\caption{\textbf{Whole-body task performance in simulation.} Direct navigation uses the standard reset without an extra camera render. All planner-navigation rows use GPT-5.6 Terra. SR and collision rate (Coll.) in \%; 15 routes per navigation setting and 32 trials for navigation--interaction, where \emph{Reach} counts arrival at the interaction region and \emph{Complete} additionally requires the object task. --: setting not applicable to, or not run for, the method.}
\label{tab:main_results}
\label{tab:vision_dependence}
\vspace{-2mm}
\footnotesize
\setlength{\tabcolsep}{3pt}
\begin{tabular*}{\textwidth}{@{\extracolsep{\fill}}l cc ccc c cc@{}}
\toprule
& \multicolumn{2}{c}{Direct navigation} & \multicolumn{3}{c}{Planner navigation} & Manipulation & \multicolumn{2}{c}{Navigation + interaction}\\
\cmidrule{2-3}\cmidrule{4-6}\cmidrule{7-7}\cmidrule{8-9}
Method & SR $\uparrow$ & SPL $\uparrow$ & SR $\uparrow$ & SPL $\uparrow$ & Coll. $\downarrow$ & SR $\uparrow$ & Reach $\uparrow$ & Complete $\uparrow$\\
\midrule
Manipulation-only specialist & -- & -- & -- & -- & -- & 32.0 & -- & -- \\
Navigation-only specialist + LAG & 73.3 & 0.574 & 46.7 & 0.330 & 40.0 & -- & -- & -- \\
GR00T~\citep{nvidia2025grootn1} & 66.7 & 0.517 & 33.3 & 0.258 & \textbf{26.7} & 33.0 & 43.8 (14/32) & 6.3 (2/32) \\
Prior-initialized reference & 73.3 & 0.540 & \textbf{53.3} & 0.385 & 46.7 & -- & -- & -- \\
HumiAct & \textbf{86.7} & \textbf{0.705} & \textbf{53.3} & \textbf{0.410} & 46.7 & \textbf{36.0} & \textbf{56.3 (18/32)} & \textbf{15.6 (5/32)} \\
\bottomrule
\end{tabular*}
\vspace{-2mm}
\end{table*}

\begin{table*}[t]
\centering
\caption{\textbf{Vision dependence on the 15 direct-navigation routes} after a camera refresh at reset. Methods share route geometry and seeds within each comparison. \textbf{(a)} SR (\%) and SPL under each input condition (seed 42). \textbf{(b)} Successful clean routes (of 15) under three inference seeds. \textbf{(c)} Mean maximum lateral deviation from the start--goal line (Lat., m) and number of stalled routes (St., progress $<0.5$\,m). \grt{} denotes the additional GR00T checkpoint and Nav.+LAG the navigation specialist.}
\label{tab:robustness_main}
\label{tab:information_removal}
\vspace{-2mm}
\footnotesize
\begin{minipage}[t]{0.605\textwidth}
\centering
\setlength{\tabcolsep}{2.6pt}\renewcommand{\arraystretch}{0.96}
\begin{tabular*}{\linewidth}[t]{@{\extracolsep{\fill}}ll cc cc cc cc@{}}
\multicolumn{10}{@{}l}{\textbf{(a) Corrupted and removed visual input}}\\[1pt]
\toprule
& & \multicolumn{2}{c}{GR00T} & \multicolumn{2}{c}{\grt} & \multicolumn{2}{c}{Nav.+LAG} & \multicolumn{2}{c}{HumiAct}\\
\cmidrule{3-4}\cmidrule{5-6}\cmidrule{7-8}\cmidrule{9-10}
& Condition & SR & SPL & SR & SPL & SR & SPL & SR & SPL\\
\midrule
& Clean & 80.0 & .606 & 86.7 & .669 & 86.7 & .674 & \textbf{93.3} &\textbf{.719}\\
\midrule
\multirow{8}{*}{\rotatebox{90}{Appearance change}}
& Defocus ($r{=}6$) & 80.0 & .634 & \textbf{86.7} & \textbf{.687} & 73.3 & .516 & 80.0 & .629\\
& Noise ($\sigma{=}40$) & \textbf{86.7} & \textbf{.683} & \textbf{86.7} & .681 & 80.0 & .586 & 80.0 & .590\\
& Brightness ($\times1.8$) & 73.3 & .550 & 66.7 & .502 & 73.3 & .529 & \textbf{86.7} & \textbf{.614}\\
& Motion blur (23\,px) & 73.3 & .556 & 80.0 & .618 & 80.0 & .577 & \textbf{86.7} & \textbf{.633}\\
& JPEG ($q{=}10$) & 73.3 & .560 & 80.0 & .626 & 80.0 & .562 & \textbf{86.7} & \textbf{.649}\\
& \emph{Mean of five} & 77.3 & .597 & 80.0 & \textbf{.623} & 77.3 & .554 & \textbf{84.0} & \textbf{.623}\\
& Strong defocus ($r{=}20$) & 46.7 & .347 & \textbf{73.3} & \textbf{.570} & \textbf{73.3} & .535 & \textbf{73.3} & .546\\
& Strong noise ($\sigma{=}100$) & 60.0 & .443 & 73.3 & \textbf{.559} & 73.3 & .521 & \textbf{80.0} & .553\\
\midrule
\multirow{2}{*}{\rotatebox{90}{Rem.}}
& Gray image & \textbf{73.3} & .495 & 66.7 & \textbf{.504} & 66.7 & .418 & 46.7 & .305\\
& Central occlusion (25\%) & \textbf{80.0} & \textbf{.574} & 66.7 & .492 & 73.3 & .479 & 60.0 & .443\\
\bottomrule
\end{tabular*}
\end{minipage}\hfill
\begin{minipage}[t]{0.375\textwidth}
\centering
\setlength{\tabcolsep}{2pt}\renewcommand{\arraystretch}{0.96}
\begin{tabular*}{\linewidth}[t]{@{\extracolsep{\fill}}l cccc@{}}
\multicolumn{5}{@{}l}{\textbf{(b) Clean routes completed per inference seed}}\\[1pt]
\toprule
Seed & GR00T & \grt & Nav.+LAG & HumiAct\\
\midrule
42 & 12 & 13 & 13 & 14\\
43 & 11 & 10 & 11 & 12\\
44 & 11 & 12 & 13 & 12\\
\emph{Mean} & 11.3 & 11.7 & 12.3 & \textbf{12.7}\\
\bottomrule
\end{tabular*}

\vspace{4pt}
\begin{tabular*}{\linewidth}[t]{@{\extracolsep{\fill}}l cc cc cc cc@{}}
\multicolumn{9}{@{}l}{\textbf{(c) Trajectory diagnostics}}\\[1pt]
\toprule
& \multicolumn{2}{c}{GR00T} & \multicolumn{2}{c}{\grt} & \multicolumn{2}{c}{Nav.+LAG} & \multicolumn{2}{c}{HumiAct}\\
\cmidrule{2-3}\cmidrule{4-5}\cmidrule{6-7}\cmidrule{8-9}
Condition & Lat. & St. & Lat. & St. & Lat. & St. & Lat. & St.\\
\midrule
Clean & .28 & 0 & .36 & 0 & .25 & 0 & .21 & 0\\
Gray image & .50 & 1 & .37 & 0 & .27 & 0 & .46 & 1\\
Occlusion & .30 & 0 & .34 & 0 & .35 & 0 & .29 & 0\\
Strong defocus & .25 & 6 & .33 & 0 & .28 & 0 & .34 & 0\\
Strong noise & .31 & 0 & .29 & 0 & .33 & 0 & .32 & 0\\
\bottomrule
\end{tabular*}
\end{minipage}
\vspace{-2mm}
\end{table*}

\subsection{Whole-Body Task Performance}
\label{sec:whole_body_results}

\begin{figure*}[t]
\centering
\begin{minipage}[b]{0.645\textwidth}
\centering
\includegraphics[width=\linewidth]{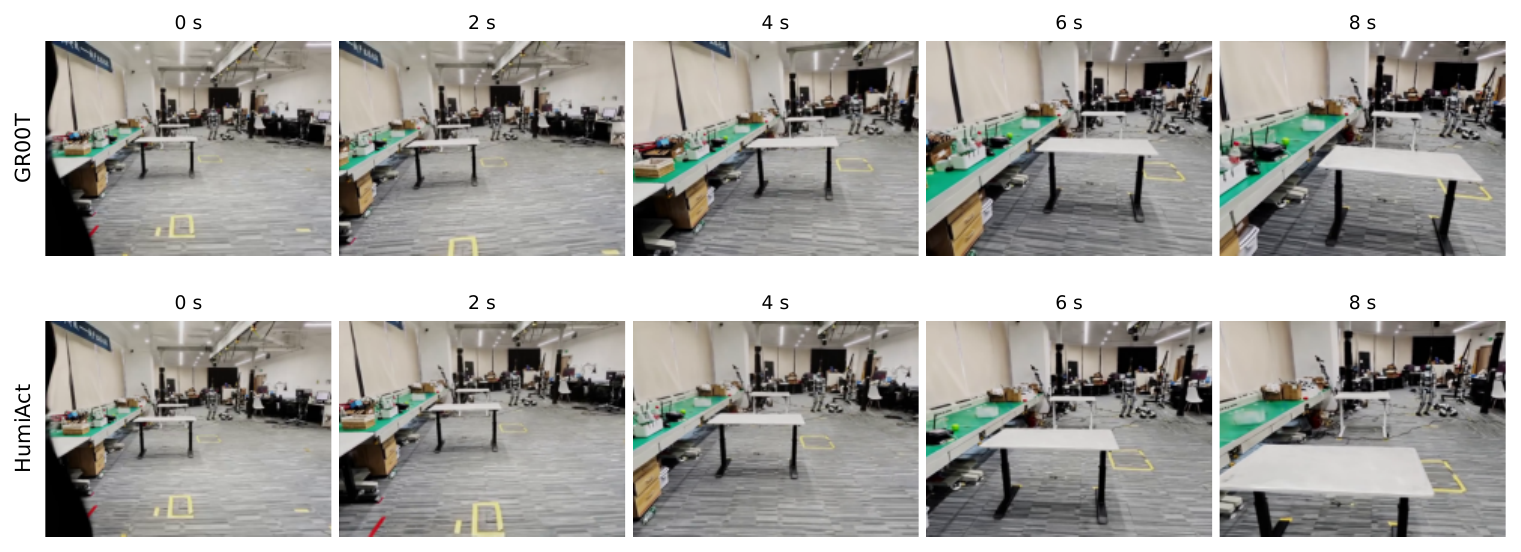}\\[-1pt]
{\footnotesize (a) Egocentric observations on the same direct-navigation route}
\end{minipage}\hfill
\begin{minipage}[b]{0.325\textwidth}
\centering
\includegraphics[width=\linewidth]{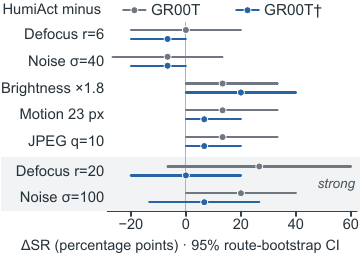}\\[-1pt]
{\footnotesize (b) Route-paired $\Delta$SR of HumiAct}
\end{minipage}
\vspace{-2mm}
\caption{\textbf{Execution and paired robustness.} \textbf{(a)} Recorded simulator frames at 0, 2, 4, 6, and 8\,s for GR00T (top) and HumiAct (bottom) on route 04. \textbf{(b)} HumiAct minus GR00T (gray) and minus \grt{} (blue) in SR points for each corruption; bars are 95\% intervals from resampling the 15 route pairs.}
\label{fig:supp_navigation_frames}
\label{fig:supp_paired_uncertainty}
\vspace{-3mm}
\end{figure*}

\paragraph{One policy for navigation and manipulation.}
Table~\ref{tab:main_results} compares policies on four whole-body tasks.
On direct routes HumiAct completes 13 of 15 (SPL 0.705) against 10 of 15 (0.517) for GR00T, gaining three routes and losing none (route-paired $\Delta$SR $+20.0$ points, 95\% bootstrap interval $[0.0, 40.0]$).
With the same planner it completes 8 obstacle routes against 5, with SPL rising from 0.258 to 0.410; Fig.~\ref{fig:supp_navigation_frames}a shows a recorded direct-navigation episode for both policies.
Sharing the policy does not trade one skill for the other: the navigation specialist with the same initialization reaches 73.3\% and 46.7\% where the joint policy reaches 86.7\% and 53.3\%, and manipulation is retained (36.0\% against 32.0\% for the manipulation specialist and 33.0\% for GR00T).
\\
\textbf{Composed navigation--interaction.}
The composed task is where a unified body matters most, since the approach determines the viewpoint and posture from which the interaction starts.
HumiAct reaches the interaction region in 18/32 trials and completes the task in 5/32, against 14/32 and 2/32 for GR00T. The navigation specialist shares HumiAct's LAG initialization, so its gap (13.4/6.6 points) isolates joint training from LAG (ablated in Table~\ref{tab:lag_ablation}).

\subsection{Vision Dependence Dilemma}
\label{sec:vision_dependence}

\begin{table}[t]
\centering
\caption{\textbf{Direct-navigation calibration} (protocol and metrics follow Table~\ref{tab:robustness_main}). With clean RGB, HumiAct reaches more goals with tighter lateral tracking than non-learned gait-replay references, including privileged-pose steering.}
\vspace{-2mm}
\label{tab:gait_reference}
\small
\setlength{\tabcolsep}{4pt}
\begin{tabular}{llccc}
\toprule
Controller & Input & SR $\uparrow$ & SPL $\uparrow$ & Lat. $\downarrow$ \\
\midrule
\multicolumn{5}{c}{\textit{Non-learned references}} \\
\midrule
Open-loop gait  & none            & 80.0 & .616 & .39 \\
Heading hold    & root yaw        & 80.0 & .617 & .39 \\
Oracle steering & true pose, goal & 86.7 & .807 & .23 \\
\midrule
\multicolumn{5}{c}{\textit{Learned policies, camera image}} 
\\\midrule
GR00T~\citep{nvidia2025grootn1} & RGB & 80.0 & .606 & .28 \\
\grt{}              & RGB & 86.7 & .669 & .36 \\
Nav.+LAG            & RGB & 86.7 & .674 & .25 \\
HumiAct             & RGB & 93.3 & .719  & .21 \\
\midrule
\multicolumn{5}{c}{\textit{Learned policies, uniform gray image}} \\
\midrule
GR00T~\citep{nvidia2025grootn1} & gray & 73.3 & .495 & .50 \\
HumiAct             & gray & 46.7 & .305 & .46 \\
\bottomrule
\end{tabular}
\vspace{-2mm}
\end{table}

\begin{figure}[h]
\centering
\vspace{-2mm}
\includegraphics[width=\linewidth]{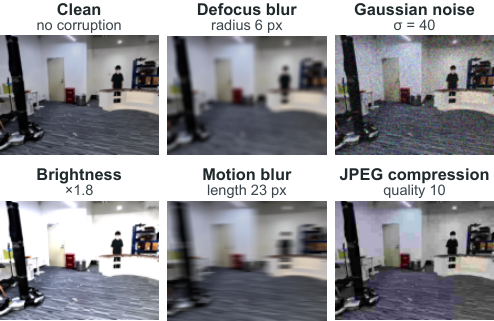}
\vspace{-2mm}
\caption{Five standard visual corruptions. Each panel shows the first policy input under the indicated corruption.}
\vspace{-2mm}
\label{fig:route08_corruptions}
\end{figure}

Following the use of appearance corruptions to probe visual
robustness~\citep{hendrycks2019corruptions}, Table~\ref{tab:robustness_main}
intervenes on the policy image alone; language, proprioception, and controller
inputs are untouched. The visual-input study renders the camera after each
reset and before the first policy action. With this protocol, HumiAct completes
14/15 clean routes, \grt{} 13/15, and GR00T 12/15, compared with 13/15 for
HumiAct and 10/15 for GR00T under the standard reset in
Table~\ref{tab:main_results}.
\\
\textbf{Calibrating direct navigation.}
We calibrate the goal-facing routes against three non-learned gait-replay references (Table~\ref{tab:gait_reference}).
Blind replay already completes most routes, making SPL and lateral drift more informative than SR alone, while yaw-only heading hold leaves drift unchanged because orientation cannot observe lateral displacement.
With camera feedback, HumiAct improves goal arrival and nearly halves lateral drift, matching privileged-pose oracle steering.
\\
\textbf{Visual feedback during walking.}
Replacing the image with uniform gray or central occlusion substantially increases lateral drift across learned policies and drops HumiAct below blind gait replay (Table~\ref{tab:robustness_main}a,c).
Rather than reverting to a nominal open-loop gait, the policy actively relies on camera feedback to stay on course and is pulled off line when visual information is removed.
\\
\textbf{Appearance corruptions.}
HumiAct leads across the five standard corruptions (84.0\% mean SR vs.\ 80.0\% for \grt{} and 77.3\% for GR00T; Table~\ref{tab:robustness_main}a, Fig.~\ref{fig:supp_paired_uncertainty}b), avoids the stalling that afflicts GR00T under strong defocus (Table~\ref{tab:robustness_main}c), and maintains the highest clean completion across inference seeds (Table~\ref{tab:robustness_main}b). Fig.~\ref{fig:route08_corruptions} shows the five inputs.
\begin{figure*}[t]
\centering
\hspace*{0.03\textwidth}\begin{minipage}[c]{0.285\textwidth}
\centering
\includegraphics[width=\linewidth]{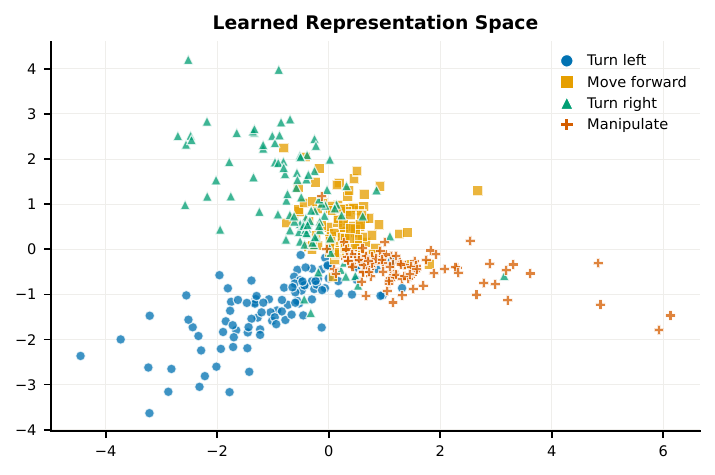}\\[-1pt]
{\footnotesize (a) Relative predictive evidence}
\end{minipage}\hfill
\begin{minipage}[c]{0.64\textwidth}
\centering
\includegraphics[width=\linewidth]{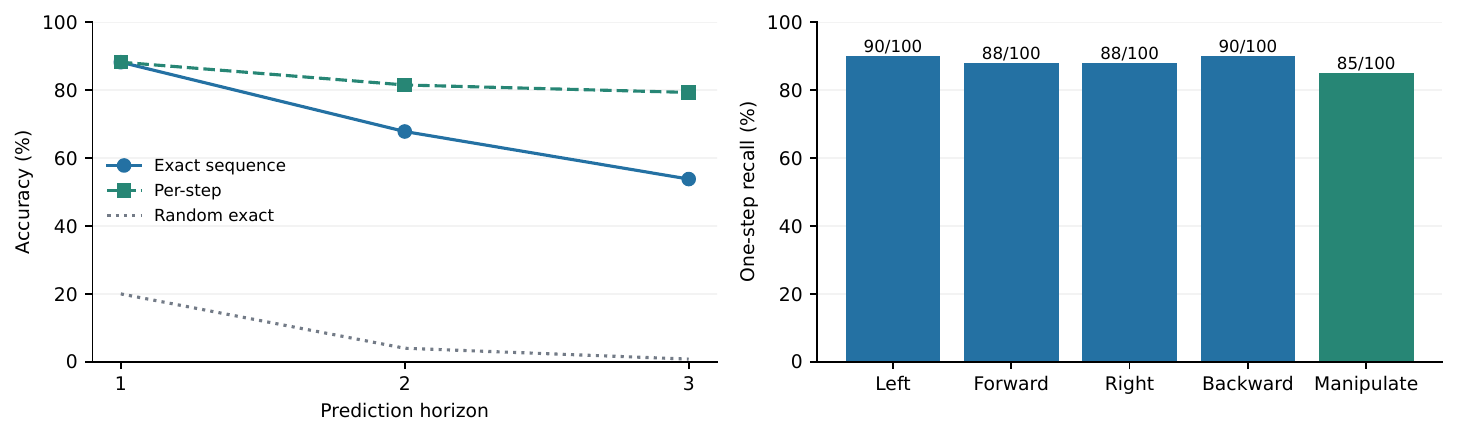}\\[-1pt]
{\footnotesize (b) Sequence accuracy over horizons \hspace{0.14\linewidth} (c) One-step recall per behavior}
\end{minipage}\hspace*{0.03\textwidth}
\vspace{-2mm}
\caption{\textbf{Learned representation space.} \textbf{(a)} Each point is a held-out clip. For every candidate behavior the predictor imagines a future, which is scored by cosine similarity to the observed future; the scores are centered within the clip, standardized, and projected by PCA (two components, 76.2\% of the variance). Clips group by their true behavior although labels are used only for coloring. \textbf{(b)} Exact-sequence and per-step accuracy against analytic chance. \textbf{(c)} One-step recall for each of the five behaviors (100 clips each).}
\label{fig:representation_pca}
\label{fig:supp_latent_horizons}
\vspace{-3mm}
\end{figure*}
\subsection{Learned Representation Space}
\label{sec:uniact_experiments}
LAG can only help if the teacher encodes how behaviors change the scene.
We test this directly on held-out clips whose source videos and action records are disjoint from training, with the predictor frozen.
\\
\textbf{Predicted futures identify the action.}
For each clip, we roll out all candidate action sequences and select the predicted future closest to the observed one.
Across the 500 balanced clips, accuracy reaches 88.2\% at $H{=}1$ and 67.8\%/53.8\% on exact two-/three-step sequences (vs.\ 4.0\%/0.8\% chance; Table~\ref{tab:uniact}), with three-step per-step accuracy near 80\% (Fig.~\ref{fig:representation_pca}b).
One-step recall spans 85--90\% across all five behaviors (Fig.~\ref{fig:representation_pca}c), confirming that minority navigation actions are represented as well as manipulation.
\begin{table}[t]
\centering
\caption{\textbf{Latent action-sequence discrimination} (\%) on 500 balanced held-out clips. \emph{Exact} / \emph{Step}: full-sequence / per-step accuracy (UniAct 95\% CI half-widths $\le 4.3$).}
\label{tab:uniact}
\vspace{-2mm}
\footnotesize
\setlength{\tabcolsep}{2pt}
\begin{tabular*}{\linewidth}{@{\extracolsep{\fill}}l c cc cc@{}}
\toprule
& $H{=}1$ & \multicolumn{2}{c}{$H{=}2$} & \multicolumn{2}{c}{$H{=}3$}\\
\cmidrule{2-2}\cmidrule{3-4}\cmidrule{5-6}
Condition & Acc. & Exact & Step & Exact & Step\\
\midrule
Random chance & 20.0 & 4.0 & 20.0 & 0.8 & 20.0\\
Train-marginal majority & 20.0 & 20.0 & 31.8 & 20.0 & 39.8\\
Permuted action labels & 3.0 & 0.7 & 4.6 & 0.3 & 5.2\\
UniAct & \textbf{88.2} & \textbf{67.8} & \textbf{81.5} & \textbf{53.8} & \textbf{79.3}\\
\bottomrule
\end{tabular*}
\vspace{-2mm}
\end{table}

\noindent\textbf{The evidence comes from the action condition.}
Permuting the action labels at inference, with the predictor unchanged, lowers one-step accuracy to 3.0\%, far below the 20\% of chance: the predictor commits to the future of the action it is given, so the wrong label systematically selects the wrong future.
The representation is therefore organized by action-conditioned dynamics and not by visual statistics of the clips.
\\
\textbf{Structure of the space.}
Fig.~\ref{fig:representation_pca}a visualizes the same evidence on the 400 clips of the four behaviors other than backward motion.
Candidate argmax selects the true behavior for 359/400 clips (89.75\%).
After removing the component shared by all candidates of a clip, two principal components explain 76.2\% of the variance of the candidate scores, and clips arrange by behavior: left and right turns extend in opposite directions along the second component, manipulation extends along the first, and forward motion lies between them.
LAG distills these action-conditioned predictive features into the policy backbone.

\subsection{Task Planning}
\label{sec:policy_planner}

\begin{table}[t]
\centering
\caption{\textbf{Planner choice} with the HumiAct policy fixed, on the 15 obstacle routes. SR and collision rate in \%.}
\label{tab:planner_main}
\label{tab:planner_coupling}
\vspace{-2mm}
\footnotesize
\setlength{\tabcolsep}{2pt}
\begin{tabular*}{0.8\linewidth}{@{\extracolsep{\fill}}lccc@{}}
\toprule
Planner & SR $\uparrow$ & SPL $\uparrow$ & Coll. $\downarrow$\\
\midrule
GPT-6 Astra & \textbf{93.3}  & \textbf{0.716} & 40.0\\
GPT-6.1 Sol & 73.3 & 0.569 & 46.7\\
GPT-5.6 Terra (default) & 53.3  & 0.410 & 46.7\\
GPT-5.6 Sol & 46.7  & 0.379 & 40.0\\
Qwen3.7-plus & 20.0  & 0.147 & 13.3\\
GPT-5.6 Luna & 6.7 & 0.041 & \textbf{6.7}\\
\bottomrule
\end{tabular*}
\vspace{-2mm}
\end{table}

With the HumiAct policy fixed, Table~\ref{tab:planner_main} varies only the planner on the obstacle routes.
SR ranges from 6.7\% to 93.3\%: GPT-6 Astra completes 14/15 routes with SPL 0.716, well above the 8/15 of the default planner used in Table~\ref{tab:main_results}.
The same motor policy therefore completes far more obstacle routes when it is given better instructions, and the planner-navigation numbers in Table~\ref{tab:main_results} are conservative with respect to the planner.
Collision rate has to be read together with SR: GPT-5.6 Luna has the fewest contacts because it rarely advances past the obstacle (1/15 routes).
\\
\textbf{Physical robot.}
The same planner--policy service and whole-body interface run on the physical humanoid (Sec.~\ref{sec:data_to_robot}).
In planner-free trials, the navigation-adapted policy achieved a higher forward-walking success rate than GR00T.
The appendix reports the trial setup and success rates for each command.

\subsection{LAG Ablation}
\label{sec:lag_ablation}
\begin{table}[t]
\centering
\caption{\textbf{LAG ablation.} Success rate (\%). Direct and Terra-guided navigation follow Table~\ref{tab:main_results}. The last column averages SR across defocus, Gaussian noise, brightness change, motion blur, and JPEG compression (Table~\ref{tab:robustness_main}a).}
\label{tab:lag_ablation}
\footnotesize
\setlength{\tabcolsep}{2pt}
\begin{tabular*}{\linewidth}{@{\extracolsep{\fill}}llrrr@{}}
\toprule
Distilled targets & Teacher action & Direct & Terra & Mean \\
\midrule
GR00T (none)    & --       & 66.7 & 33.3 & 77.3 \\
Current         & Aligned  & 73.3 & \textbf{53.3} & 78.7 \\
Future          & None     & 66.7 & 46.7 & 76.0 \\
Current + future & None     & 73.3 & 46.7 & 80.0 \\
Current + future & Shuffled & 46.7 & 33.3 & 74.6 \\
HumiAct (both)   & Aligned  & \textbf{86.7} & \textbf{53.3} & \textbf{84.0} \\
\bottomrule
\end{tabular*}
\end{table}

With both feature targets, direct-navigation SR is 86.7\% with aligned teacher actions, 73.3\% without action conditioning, and 46.7\% with shuffled actions (Table~\ref{tab:lag_ablation}). The current-only variant matches HumiAct on Terra-guided navigation (53.3\%) and averages 78.7\% across the five corruptions, compared with 84.0\% for HumiAct.

\section{Conclusion}
\label{sec:conclusion}

HumiAct takes a step toward unified humanoid navigation and manipulation by using one whole-body policy throughout the approach to an object and the ensuing interaction. During training, latent action guidance distills action-conditioned visual predictions from cross-embodiment trajectories; at inference, the policy runs without the predictor. In simulation, HumiAct reached the interaction region more often and completed more navigation--interaction trials than GR00T. On the real humanoid, HumiAct demonstrated coordinated stepping and pushing during chair repositioning.

Whole-body navigation and manipulation is challenging because the approach determines the stance, heading, and viewpoint for the subsequent interaction, so locomotion errors compound into manipulation failures. This coupling motivates a single policy that handles both phases without switching controllers. Future work includes controlled comparisons with modular expert pipelines, evaluation in unseen scenes, and broader real-robot tasks requiring concurrent locomotion and manipulation.
\clearpage
% Restore the full column height after flushing two-column floats.
\makeatletter
\global\@colroom=\textheight
\global\vsize=\textheight
\makeatother

{
\small
% Self-contained BibTeX-compiled bibliography (generated from main_references.bib with ieeenat_fullname.bst).

}

\clearpage
\makeatletter\humi@appendixtrue\makeatother
\maketitlesupplementary
\appendix
\setcounter{figure}{0}
\renewcommand{\thefigure}{S\arabic{figure}}
\setcounter{table}{0}
\renewcommand{\thetable}{S\arabic{table}}
\setlist[enumerate]{topsep=2pt,itemsep=2pt,parsep=0pt,leftmargin=1.4em}
\setlist[itemize]{topsep=2pt,itemsep=1.5pt,parsep=0pt,leftmargin=1.2em}
\renewcommand{\topfraction}{0.88}
\renewcommand{\dbltopfraction}{0.88}
\renewcommand{\textfraction}{0.10}
\renewcommand{\floatpagefraction}{0.75}
\renewcommand{\dblfloatpagefraction}{0.75}
\setlength{\dbltextfloatsep}{6pt plus 2pt minus 2pt}
\setlength{\dblfloatsep}{5pt plus 2pt minus 2pt}
\setlength{\textfloatsep}{6pt plus 2pt minus 2pt}
\makeatletter
\setlength{\@dblfptop}{0pt}
\setlength{\@dblfpsep}{12pt}
\setlength{\@dblfpbot}{0pt plus 1fil}
\renewcommand\paragraph{\@startsection{paragraph}{4}{\z@}{1.1ex \@plus0.4ex \@minus0.2ex}{-0.8em}{\normalfont\normalsize\bfseries}}
\makeatother
\raggedbottom

This appendix provides reproducibility details, route-paired statistical tests, root-cause analyses of anomalous metrics and failure modes, trajectory and corruption visualizations, UniAct/LAG ablation specifications, physical-robot trial details, and a summary of qualitative simulation and real-humanoid rollouts. Every metric, hyperparameter, and symbol follows the main paper.

% ==========================================================================
\section{Reproducibility, Compute, and Evaluation Protocols}
\label{app:protocol}
% ==========================================================================

\noindent\textbf{Claim anchor.}
\emph{Supports Main Sec.~\mainref{sec:data_to_robot} and Sec.~\mainref{sec:experimental_setup} (Tables~\mainref{tab:training_data}--\mainref{tab:robustness_main}).}
\emph{Takeaway:} All learned policies share the same humanoid corpus, downstream fine-tuning budget, and whole-body controller with zero inference-time overhead from LAG, while paired evaluation protocols maximize statistical resolution under hardware-aligned trial budgets.

\begin{table*}[t]
\centering
\caption{\textbf{Reproducibility, compute, and baseline-fairness summary} across all three training stages and evaluation protocols. Every hyperparameter, episode count, and protocol parameter matches Main Sec.~\mainref{sec:data_to_robot} and Sec.~\mainref{sec:experimental_setup}. Teacher weights, auxiliary tokens, poolers ($Q_{\mathrm{cur}}, Q_{\mathrm{fut}}$), and projector $G$ are discarded after Stage~2, incurring zero inference overhead.}
\label{tab:supp_repro}
\vspace{-2mm}
\footnotesize
\setlength{\tabcolsep}{3.5pt}\renewcommand{\arraystretch}{0.95}
\begin{tabular*}{\textwidth}{@{\extracolsep{\fill}}p{.17\textwidth}p{.26\textwidth}p{.27\textwidth}p{.25\textwidth}@{}}
\toprule
Component & Stage 1: V-JEPA-AC Prior & Stage 2: LAG Alignment & Stage 3: Whole-Body Fine-Tuning \\
\midrule
Base / frozen modules & Frozen ViT-giant $E_\phi$ ($384{\times}384$, patch~16, 576 tokens, width 1408) & Frozen teacher $(E_\phi, P_\psi)$; GR00T N1 base; 32 aux.\ tokens (16 cur + 16 fut) & LAG-aligned backbone + action modules (GR00T/\grt{} start from public base) \\
Trainable modules & 24-layer causal $P_\psi$ (width 1024, 16 heads, RoPE, 7-D zero-padded $\mathbf{u}_t$) & Selected LLM layers (3rd-from-last, width 2048), shared $G$, poolers $Q_{\mathrm{cur}}, Q_{\mathrm{fut}}$ & Full VLA backbone, state/action projections, flow-matching action transformer $f_\theta$ \\
Training corpus & 178,384 episodes: AgiBotWorld (142,069), Galaxea (34,353), Humanoid (1,962) & Same 178,384-episode heterogeneous corpus (video-/action-/parquet-disjoint split) & 5,421 humanoid whole-body demos across 84 directories (49 nav./manip.\ + 35 converted) \\
Objective & $\mathcal{L}_{\mathrm{AC}}$: normalized $\ell_1$ (teacher-forced + $K{=}2$ AR steps; 4 frames @ 2\,Hz) & $\mathcal{L}_{\mathrm{LAG}}$: token-wise cosine distance ($\lambda_{\mathrm{cur}}{=}\lambda_{\mathrm{fut}}{=}1.0$; 576$\to$16 pooled) & $\mathcal{L}_{\mathrm{FM}}$: masked flow matching; 40-step chunk (30 exec.\ @ 20\,Hz sim; 30 @ 10\,Hz real) \\
Optimization & 5,000 iters (10 ep.\ $\times$ 500), peak LR $1.5{\times}10^{-4}\!\to\!10^{-6}$, 2-ep.\ warmup, wd 0.04 & LR $10^{-4}$, wd $3{\times}10^{-5}$, warmup 5\%, color jitter; discarded before inference & 200k steps, batch 96, LR $10^{-4}$, wd $10^{-5}$, warmup 5\%, state dropout 0.2 (shared by GR00T/\grt) \\
Compute \& precision & 16 NVIDIA H20 GPUs, BF16; action weights $[2,2,2,1,2]$ & 16 NVIDIA H20 GPUs, BF16 & 16 NVIDIA H20 GPUs, BF16; deploy on RTX 5000 Ada (4 flow steps, 10/50\,Hz) \\
Seeds \& tuning budget & Seed 20260916; final ckpt.\ (5k iters); PCA seed 20260923; no test tuning & Inherited recipe; identical downstream budget for all LAG ablations (Table~\mainref{tab:lag_ablation}) & Eval seeds 42--44 ($42{+}10{,}000e{+}q$); bootstrap 10,000 draws (seed 20261111) \\
\bottomrule
\end{tabular*}
\end{table*}

\paragraph{Baseline Fairness and Tuning Budget.}
Table~\ref{tab:supp_repro} summarizes architecture, data, optimization, compute, and seed configurations across all three training stages.
All compared whole-body policies---GR00T~\citep{nvidia2025grootn1}, \grt, the navigation-only specialist (Nav.+LAG), the manipulation-only specialist, and HumiAct---share the same 84-directory whole-body demonstration corpus (5,421 episodes: 49 original navigation/manipulation directories totaling 2,205 episodes plus 35 converted directories totaling 3,216 episodes), the same 200k-step fine-tuning schedule (global batch size 96, learning rate $10^{-4}$, weight decay $10^{-5}$, 5\% linear warmup, state dropout 0.2), the same flow-matching action transformer $f_\theta$~\citep{lipman2023flow}, and the same 50\,Hz BeyondMimic whole-body tracker~\citep{liao2025beyondmimic}.
\grt{} additionally matches HumiAct's exact downstream data order and optimization recipe from the public GR00T base model.
All hyperparameters are inherited from the base GR00T and V-JEPA~2~\citep{assran2025vjepa2} recipes without per-method tuning on evaluation routes.
Because the V-JEPA-AC teacher $(E_\phi, P_\psi)$, the 32 auxiliary alignment tokens, the poolers $(Q_{\mathrm{cur}}, Q_{\mathrm{fut}})$, and the projector $G$ are discarded prior to deployment, HumiAct incurs zero inference latency or parameter overhead relative to GR00T.

\paragraph{Hardware-Aligned Trial Budgets and Paired Statistical Design.}
On a full-size 33-DoF physical humanoid, each closed-loop trial requires full workspace reset, battery and thermal monitoring, safety spotting, and balance stabilization.
We intentionally designed the evaluation suite from the outset with a compact, hardware-feasible trial structure ($N{=}15$ routes for direct and planner navigation, $N{=}32$ starts for composed navigation--interaction, and $N{=}3$ repetitions per primitive on hardware) so that simulation and physical-robot evaluation remain structurally aligned under one shared control interface.
To maintain statistical rigor under compact $N$, all simulation model comparisons use \emph{route-paired evaluation}: compared checkpoints share the exact start pose, goal position, obstacle layout, environment seed, and per-request policy noise seed $42{+}10{,}000e{+}q$ for request $q$ of episode $e$.
Gaussian image noise draws from a separate deterministic stream indexed by episode, request, batch, and step.
We quantify uncertainty via 10,000-draw route-paired percentile bootstrap 95\% confidence intervals~\citep{efron1993bootstrap} (seed 20261111), exact two-sided McNemar tests~\citep{mcnemar1947note} on discordant success pairs, paired Wilcoxon signed-rank tests~\citep{wilcoxon1945individual} on continuous SPL~\citep{anderson2018evaluation}, Fisher exact tests~\citep{fisher1922interpretation} on unpaired composed counts, and three-seed clean repeats (seeds 42--44, 45 rollouts/policy).

\paragraph{Metrics and Protocol Separation.}
Across $N$ trials with binary goal indicators $S_i \in \{0,1\}$, positive reference start--goal distance $d_i > 0$ ($2.20 \le d_i \le 4.30$\,m on direct navigation), and traveled path length $L_i$, we compute
\begin{equation}
\mathrm{SR}=\frac{1}{N}\sum_{i=1}^{N}S_i,
\quad
\mathrm{SPL}=\frac{1}{N}\sum_{i=1}^{N}S_i\,\frac{d_i}{\max(d_i,L_i)}.
\label{eq:supp_sr_spl}
\end{equation}
On direct navigation (goal-aligned start, forward-motion instruction, 10\,s horizon, 0.5\,m goal radius, no obstacles), $L_i$ spans the full 10\,s episode and $d_i$ is Euclidean start--goal distance; a route can therefore enter the goal radius and overshoot slightly afterward.
On planner navigation (15 obstacle routes), $L_i$ terminates at first goal arrival and obstacle collision rate (Coll.) records the marginal fraction of trials with any contact exceeding 10\,N; a trial may incur a light contact and still reach the goal.
On manipulation only, mean successful control steps are averaged over completed trials alongside SR.
On composed navigation--interaction ($N{=}32$), letting $A_i, B_i \in \{0,1\}$ denote interaction-region arrival and object-task completion, we report Reach ($N^{-1}\sum_i A_i$) and Complete ($N^{-1}\sum_i A_i B_i$).
In simulation, policies predict 40-step chunks at 20\,Hz and execute the first 30 steps (1.5\,s) with zero overlap.

Crucially, the whole-body comparison (Main Table~\mainref{tab:main_results}) and the visual-intervention suite (Main Tables~\mainref{tab:robustness_main}--\mainref{tab:gait_reference}) use two distinct simulator reset entry points that are never cross-compared: Table~\mainref{tab:main_results} uses the standard environment reset, whereas Tables~\mainref{tab:robustness_main}--\mainref{tab:gait_reference} execute two extra kinematic/camera render passes immediately after reset (without advancing simulation time or controller history) so the first policy frame reflects the freshly reset pose before image corruption is injected.

% ==========================================================================
\section{Statistical Analysis, Trajectory Atlas, and Visual Corruptions}
\label{app:robustness}
% ==========================================================================

\noindent\textbf{Claim anchor.}
\emph{Supports Main Sec.~\mainref{sec:whole_body_results}--\mainref{sec:vision_dependence} (Tables~\mainref{tab:main_results}--\mainref{tab:gait_reference}, Figs.~\mainref{fig:supp_navigation_frames}--\mainref{fig:route08_corruptions}).}
\emph{Takeaway:} Route-paired intervals confirm systematic SPL gains on direct navigation, statistically significant elimination of strong-defocus stalling vs.\ GR00T ($p{=}0.031$), tighter corridor tracking ($1/15$ vs.\ $4/15$ excursions $>0.5$\,m), and statistically significant degradation under gray-image removal ($p{=}0.041$), alongside exact pixel-space corruption definitions.

\begin{table*}[t]
\centering
\caption{\textbf{Route-paired statistical analysis and multi-seed clean variance.} \textbf{(a)} Paired SR ($\Delta$SR, pp) and SPL ($\Delta$SPL) differences with 95\% route-bootstrap CIs (10,000 draws, seed 20261111), gained/lost route counts (G/L), exact two-sided McNemar $p$-values (pooled over 75 paired route--condition trials for the five-corruption mean), and paired Wilcoxon signed-rank $p$-values on SPL. Bold CIs exclude zero. Under strong defocus, GR00T stalls on 6/15 routes vs.\ 0/15 for HumiAct (exact McNemar $p{=}0.031$). \textbf{(b)} Clean direct-navigation completions (out of 15) and SPL across inference seeds 42--44 (45 rollouts/policy), matching Main Table~\mainref{tab:robustness_main}b.}
\label{tab:supp_paired_stats}
\vspace{-2mm}
\footnotesize
\setlength{\tabcolsep}{3.5pt}\renewcommand{\arraystretch}{0.94}
\begin{tabular*}{\textwidth}{@{\extracolsep{\fill}}l rccr rcr@{}}
\multicolumn{8}{@{}l}{\textbf{(a) Route-paired hypothesis tests and 95\% bootstrap confidence intervals}}\\[1pt]
\toprule
Comparison & $\Delta$SR & 95\% CI & G/L & $p_{\mathrm{McN}}$ & $\Delta$SPL & 95\% CI & $p_{\mathrm{Wilc}}$ \\
\midrule
\multicolumn{8}{@{}l}{\emph{Direct navigation, standard reset (Main Table~\mainref{tab:main_results}; 15 paired routes)}} \\
HumiAct $-$ GR00T & +20.0 & [0.0, 40.0] & 3/0 & 0.250 & +0.188 & \textbf{[0.016, 0.391]} & 0.136 \\
HumiAct $-$ Prior-init.\ ref. & +13.3 & [0.0, 33.3] & 2/0 & 0.500 & +0.165 & \textbf{[0.014, 0.349]} & 0.221 \\
\midrule
\multicolumn{8}{@{}l}{\emph{Five-corruption mean, reset-refreshed (Main Table~\mainref{tab:robustness_main}a; 15 routes $\times$ 5 corruptions = 75 paired trials)}} \\
HumiAct $-$ GR00T & +6.7 & [$-$6.7, 24.0] & 8/3 & 0.227 & +0.026 & [$-$0.094, 0.168] & 0.594 \\
HumiAct $-$ \grt & +4.0 & [$-$2.7, 10.7] & 5/2 & 0.453 & +0.000 & [$-$0.057, 0.060] & 0.917 \\
HumiAct $-$ Nav.+LAG & +6.7 & [$-$1.3, 16.0] & 7/2 & 0.180 & +0.069 & \textbf{[0.004, 0.142]} & 0.087 \\
\midrule
\multicolumn{8}{@{}l}{\emph{Strong corruptions and visual-information removal, reset-refreshed (seed 42; 15 paired routes)}} \\
Strong defocus ($r{=}20$) vs.\ GR00T & +26.7 & [$-$6.7, 60.0] & 6/2 & 0.289 & +0.199 & [$-$0.092, 0.475] & 0.311 \\
Strong defocus ($r{=}20$) vs.\ \grt & +0.0 & [$-$20.0, 20.0] & 1/1 & 1.000 & $-$0.024 & [$-$0.213, 0.165] & 0.433 \\
Strong noise ($\sigma{=}100$) vs.\ GR00T & +20.0 & [0.0, 40.0] & 3/0 & 0.250 & +0.110 & [$-$0.035, 0.282] & 0.695 \\
Gray image vs.\ GR00T & $-$26.7 & \textbf{[$-$46.7, $-$6.7]} & 0/4 & 0.125 & $-$0.190 & \textbf{[$-$0.351, $-$0.048]} & 0.041 \\
Central occlusion vs.\ GR00T & $-$20.0 & [$-$46.7, 6.7] & 1/4 & 0.375 & $-$0.131 & [$-$0.347, 0.084] & 0.422 \\
\bottomrule
\end{tabular*}
\vspace{5pt}
\begin{tabular*}{\textwidth}{@{\extracolsep{\fill}}l cccc cc@{}}
\multicolumn{7}{@{}l}{\textbf{(b) Clean reset-refreshed direct navigation across inference seeds (completions / 15 and SPL)}}\\[1pt]
\toprule
Policy & Seed 42 & Seed 43 & Seed 44 & Mean $\pm$ SD (/15) & Mean SR $\pm$ SD (\%) & Mean SPL $\pm$ SD \\
\midrule
GR00T & 12 (0.606) & 11 (0.583) & 11 (0.549) & 11.3 $\pm$ 0.6 & 75.6 $\pm$ 3.8 & 0.579 $\pm$ 0.029 \\
\grt & 13 (0.669) & 10 (0.520) & 12 (0.609) & 11.7 $\pm$ 1.5 & 77.8 $\pm$ 10.2 & 0.599 $\pm$ 0.075 \\
Nav.+LAG & 13 (0.674) & 11 (0.576) & 13 (0.641) & 12.3 $\pm$ 1.2 & 82.2 $\pm$ 7.7 & 0.630 $\pm$ 0.050 \\
HumiAct & 14 (0.719) & 12 (0.612) & 12 (0.605) & 12.7 $\pm$ 1.2 & 84.4 $\pm$ 7.7 & 0.645 $\pm$ 0.064 \\
\bottomrule
\end{tabular*}
\end{table*}

\begin{table*}[t]
\centering
\caption{\textbf{Task decomposition, manipulation retention, and corridor excursion} (standard whole-body protocol, supporting Main Sec.~\mainref{sec:whole_body_results} and Table~\mainref{tab:main_results}). \textbf{(a)} Mutually exclusive outcomes on the 32 composed navigation--interaction trials; end-to-end gain is $3/32 = +9.4$\,pp (two-sided Fisher exact: Reach $p{=}0.454$, Full $p{=}0.426$, Complete\,$\mid$\,Reach $p{=}0.426$). \textbf{(b)} Manipulation-only evaluation (mean steps over successful trials; GR00T achieves 33.0\% SR in Main Table~\mainref{tab:main_results}, and the separate real-robot manipulation baseline reaches 35.0\%). \textbf{(c)} Pre-arrival lateral corridor excursion $E_i$ (Eq.~\ref{eq:supp_excursion}) across the 15 direct-navigation routes.}
\label{tab:supp_task_decomp}
\vspace{-2mm}
\footnotesize
\setlength{\tabcolsep}{3pt}\renewcommand{\arraystretch}{0.95}
\begin{minipage}[t]{0.42\textwidth}
\centering
\begin{tabular*}{\linewidth}{@{\extracolsep{\fill}}lcccc@{\extracolsep{0pt}}}
\multicolumn{5}{@{}l}{\textbf{(a) Composed task ($N{=}32$)}}\\[1pt]
\toprule
Method & No arr. & Arr.\ only & Full & Full\,$\mid$\,Arr. \\
\midrule
GR00T & 18 & 12 & 2 & 14.3\% (2/14) \\
HumiAct & 14 & 13 & 5 & 27.8\% (5/18) \\
\bottomrule
\end{tabular*}
\end{minipage}\hfill
\begin{minipage}[t]{0.25\textwidth}
\centering
\begin{tabular*}{\linewidth}{@{\extracolsep{\fill}}lrr@{\extracolsep{0pt}}}
\multicolumn{3}{@{}l}{\textbf{(b) Manipulation}}\\[1pt]
\toprule
Policy & SR (\%) & Steps \\
\midrule
Manip.-only & 32.0 & 551 \\
HumiAct & 36.0 & 566 \\
\bottomrule
\end{tabular*}
\end{minipage}\hfill
\begin{minipage}[t]{0.30\textwidth}
\centering
\begin{tabular*}{\linewidth}{@{\extracolsep{\fill}}lccc@{\extracolsep{0pt}}}
\multicolumn{4}{@{}l}{\textbf{(c) Direct-nav.\ excursion}}\\[1pt]
\toprule
Policy & Succ. & $E_i\!>\!0.5$\,m & Fall \\
\midrule
GR00T & 10/15 & 4/15 & 1/15 \\
Prior ref. & 11/15 & 0/15 & 1/15 \\
HumiAct & 13/15 & 1/15 & 1/15 \\
\bottomrule
\end{tabular*}
\end{minipage}
\end{table*}

\begin{figure*}[t]
\centering
\includegraphics[width=\textwidth]{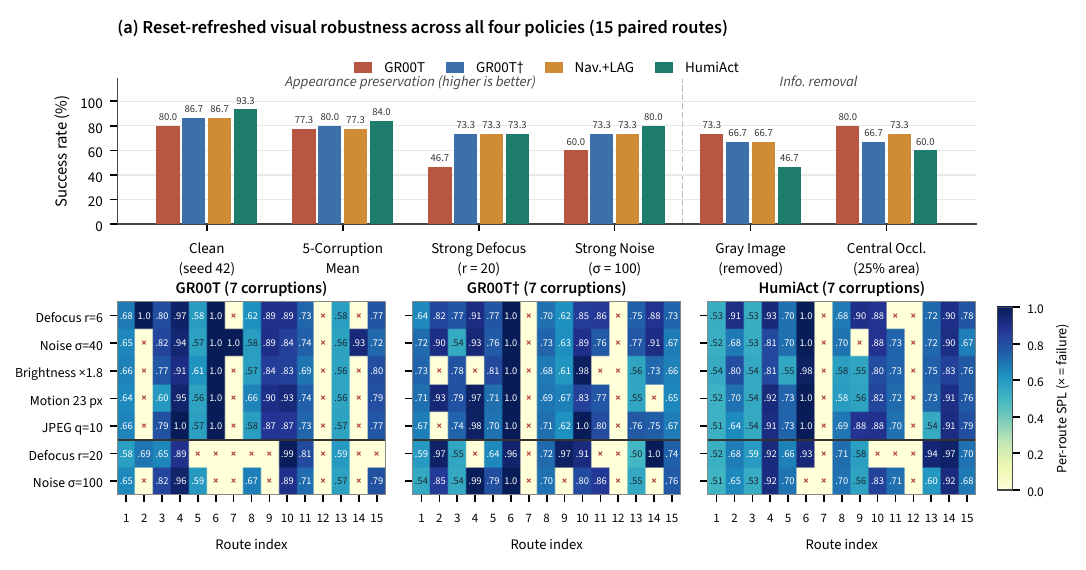}
\vspace{-5mm}
\caption{\textbf{Four-policy visual robustness and complete per-route corruption breakdown} (reset-refreshed protocol, seed 42, 15 paired routes). \textbf{(a)} Success rate across all four policies on clean input, the five-corruption mean, stronger corruptions, and visual-information removal (matching Main Table~\mainref{tab:robustness_main}a). \textbf{(b)} Complete route-by-condition SPL heatmaps across all seven appearance corruptions for GR00T, \grt, and HumiAct ($\times$ denotes non-arrival). No route is omitted.}
\label{fig:supp_rob_4m}
\vspace{-2mm}
\end{figure*}

\paragraph{Paired Hypothesis Tests and Multi-Seed Variance.}
Table~\ref{tab:supp_paired_stats}a reports paired $\Delta$SR, $\Delta$SPL, 95\% bootstrap intervals, discordant pair counts (gained/lost), and exact non-parametric tests.
On standard-reset direct navigation (Main Table~\mainref{tab:main_results}), HumiAct gains 3 routes over GR00T and 2 over the prior-initialized reference while losing none, with strictly positive 95\% bootstrap CIs on $\Delta$SPL ($+0.188\,[+0.016, +0.391]$ vs.\ GR00T; $+0.165\,[+0.014, +0.349]$ vs.\ prior reference).
Across the five standard corruptions (75 paired route--condition trials), HumiAct leads all baselines in mean SR ($84.0\%$ vs.\ $80.0\%$ for \grt{} and $77.3\%$ for GR00T and Nav.+LAG), with a strictly positive 95\% $\Delta$SPL interval over Nav.+LAG ($+0.069\,[+0.004, +0.142]$), isolating the benefit of joint navigation--manipulation training given identical LAG initialization.
Across inference seeds 42--44 on clean reset-refreshed routes (Table~\ref{tab:supp_paired_stats}b; 45 rollouts/policy), HumiAct achieves the highest mean completion ($12.7 \pm 1.2 / 15$, $84.4 \pm 7.7\%$) and mean SPL ($0.645 \pm 0.064$).
Figure~\ref{fig:supp_rob_4m} visualizes all four policies alongside the complete route-level SPL matrix across all seven corruption conditions.

\paragraph{Task Decomposition and Manipulation Retention.}
Table~\ref{tab:supp_task_decomp}a decomposes the 32 composed navigation--interaction trials into mutually exclusive outcomes: HumiAct increases interaction-region arrival by $+12.5$\,pp ($18/32$ vs.\ $14/32$) and full end-to-end completion by $+9.4$\,pp ($5/32 - 2/32 = 3/32$).
Among trials that reach the interaction region, HumiAct nearly doubles conditional object completion ($5/18 = 27.8\%$ vs.\ $2/14 = 14.3\%$), showing that whole-body approach quality carries directly into manipulation readiness.
On manipulation-only evaluation (Table~\ref{tab:supp_task_decomp}b), joint training retains manipulation accuracy ($36.0\%$ vs.\ $32.0\%$ for the specialist and $33.0\%$ for GR00T in Main Table~\mainref{tab:main_results}) with only a 15-step difference in mean completion length ($566$ vs.\ $551$ steps; on hardware, the separate manipulation-only baseline reaches $35.0\%$).

\begin{figure*}[t]
\centering
\includegraphics[width=0.95\textwidth]{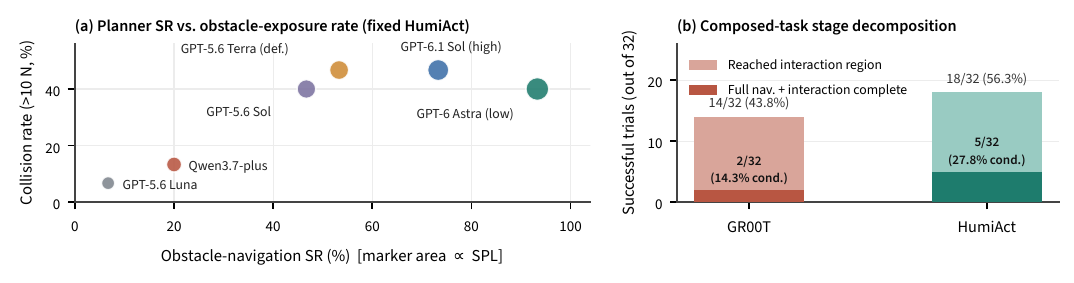}
\vspace{-4mm}
\caption{\textbf{Planner exposure trade-off and composed-task funnel.} \textbf{(a)} Obstacle-navigation SR vs.\ collision rate ($>10$\,N) across six planners with the HumiAct motor policy held fixed (marker area $\propto$ SPL; Main Table~\mainref{tab:planner_main}). \textbf{(b)} Composed-task arrival and conditional completion counts on 32 trials (Main Table~\mainref{tab:main_results}).}
\label{fig:supp_planner_funnel}
\vspace{-2mm}
\end{figure*}

\paragraph{Root-Cause Analysis of Anomalous Metrics and Failure Modes.}
Several non-monotonic numbers in the main tables warrant explicit mechanistic explanation so they are not misread as contradictions:
\begin{enumerate}
\item \emph{Larger SR drop of HumiAct under gray image and central occlusion (46.7\% vs.\ 73.3\% for GR00T; Table~\mainref{tab:robustness_main}a):}
Under a uniform gray image, HumiAct suffers a statistically significant drop relative to GR00T ($\Delta\mathrm{SR} = -26.7$\,pp, 95\% CI $[-46.7, -6.7]$; $\Delta\mathrm{SPL}$ Wilcoxon $p{=}0.041$) while its lateral deviation more than doubles ($0.21 \to 0.46$\,m).
This is expected behavior of closed-loop visual control, not fragility: because the direct-navigation routes start goal-aligned, even a blind open-loop gait replay achieves $80.0\%$ SR with $0.39$\,m drift (Main Table~\mainref{tab:gait_reference}).
A policy that falls back on proprioceptive open-loop walking when images vanish retains high SR on straight routes, whereas HumiAct actively steers via visual feedback (achieving $0.21$\,m clean drift, matching the $0.23$\,m privileged oracle) and is therefore pulled off line when visual information is stripped completely.

\item \emph{Single-condition fluctuations under moderate noise ($\sigma{=}40$: 80.0\% vs.\ 86.7\%) and defocus ($r{=}6$: 80.0\% vs.\ 86.7\% for \grt):}
On $N{=}15$ routes, one route equals $6.7$\,pp. Figure~\ref{fig:supp_rob_4m}b confirms that both gaps stem from a single route flip (Route 9 under $\sigma{=}40$; Route 11 under $r{=}6$; exact McNemar $p{=}1.0$), whereas HumiAct wins on Brightness ($+13.3/+20.0$\,pp), Motion blur ($+13.3/+6.7$\,pp), JPEG ($+13.3/+6.7$\,pp), Strong noise ($+20.0/+6.7$\,pp), and Clean ($+13.3/+6.7$\,pp).
Under strong defocus ($r{=}20$), GR00T stalls (progress $<0.5$\,m) on $6/15$ routes vs.\ $0/15$ for HumiAct (exact McNemar $p{=}0.031$).
Notably, \grt{} also avoids stalling ($73.3\%$, 0 stalls) and shows a smaller relative SR drop on the two strong corruptions ($15.4\%$ vs.\ $17.9\%$ for HumiAct and $33.3\%$ for GR00T, computed via $D_{m,c}=(\mathrm{SR}_{m,\mathrm{clean}}-\mathrm{SR}_{m,c})/\mathrm{SR}_{m,\mathrm{clean}}$).
Thus, matching HumiAct's downstream training recipe largely eliminates catastrophic freezing, while LAG provides additional gains in clean lateral tracking ($0.21$\,m vs.\ $0.36$\,m), five-corruption mean SR ($+4.0$\,pp), and strong noise ($+6.7$\,pp).

\item \emph{Higher collision rate accompanying higher SR on planner navigation (46.7\% vs.\ 26.7\% for GR00T; 40.0\% for GPT-6 Astra vs.\ 6.7\% for GPT-5.6 Luna):}
Collision rate is a marginal count of $>10$\,N contact events across 15 routes and is coupled to obstacle exposure.
Policies or planners that stall or turn away before reaching the obstacle rarely trigger contact (e.g., GPT-5.6 Luna completes $1/15$ routes with $1/15$ collisions), whereas trajectories that actively navigate past narrow clearances accumulate brief contacts even while reaching the goal.
Consequently, SR, SPL, and Coll.\ must always be interpreted jointly (Fig.~\ref{fig:supp_planner_funnel}a).

\item \emph{Non-learned gait-replay calibration (80.0\% SR on goal-facing routes):}
The three non-learned references in Main Table~\mainref{tab:gait_reference} loop one steady 33-step (1.65\,s) forward-walking gait cycle resampled to 20\,Hz through the same whole-body interface.
Open-loop gait and IMU-style heading hold both reach $12/15$ goals and fail solely on the three longest routes (3.49, 4.25, and 4.30\,m), which exceed their 10\,s travel range; privileged oracle steering (replanning every 0.5\,s from ground-truth pose) misses only the two longest (4.25 and 4.30\,m).
This calibration clarifies scope: goal-facing straight routes primarily test closed-loop lateral tracking (where HumiAct cuts drift nearly in half, $0.39 \to 0.21$\,m) and visual robustness, whereas obstacle routes and composed tasks require closed-loop perception, planning, and loco-manipulation that gait replay cannot perform.

\item \emph{Modest absolute completion on composed navigation--interaction (5/32 = 15.6\%):}
End-to-end loco-manipulation compounds errors across long horizons: small heading or foot-placement deviations at arrival shift the object out of the reliable grasp envelope (13/18 arrivals fail during interaction).
This reflects genuine task difficulty rather than policy collapse, as confirmed by the nearly $2\times$ higher conditional completion rate over GR00T ($27.8\%$ vs.\ $14.3\%$), and identifies close-range stance alignment as a key direction for future work.
\end{enumerate}

\begin{figure*}[t]
\centering
\includegraphics[width=\textwidth]{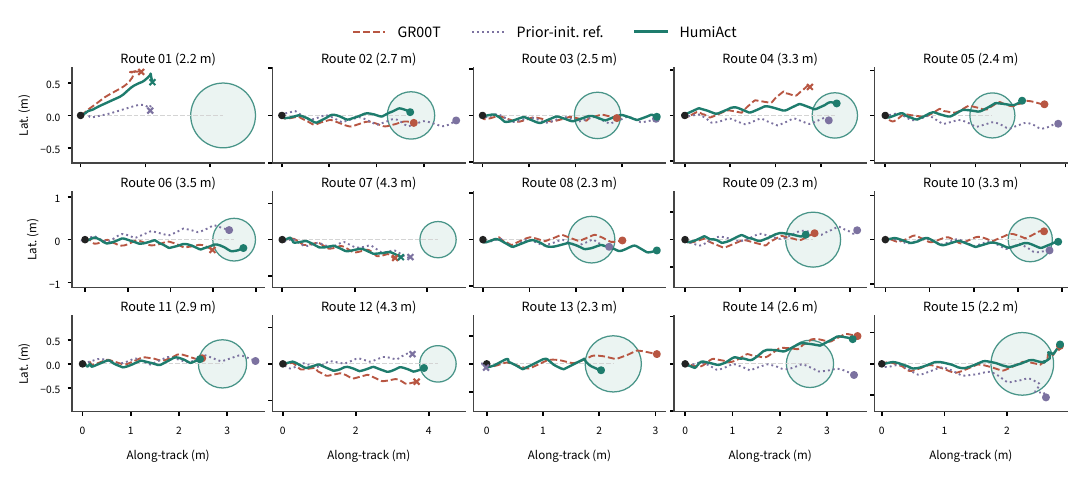}
\vspace{-5mm}
\caption{\textbf{Complete 15-route direct-navigation trajectory atlas} (standard whole-body protocol, Main Table~\mainref{tab:main_results}). Each route $i$ is translated to $\mathbf{p}_i^{\mathrm{start}}$ and rotated so $\mathbf{p}_i^{\mathrm{goal}}$ lies on the positive horizontal axis; shaded circles mark the 0.5\,m success radius, filled dots mark completion, and $\times$ marks non-arrival.}
\label{fig:supp_route_atlas}
\vspace{-2mm}
\end{figure*}

\begin{figure*}[t]
\centering
\includegraphics[width=\textwidth]{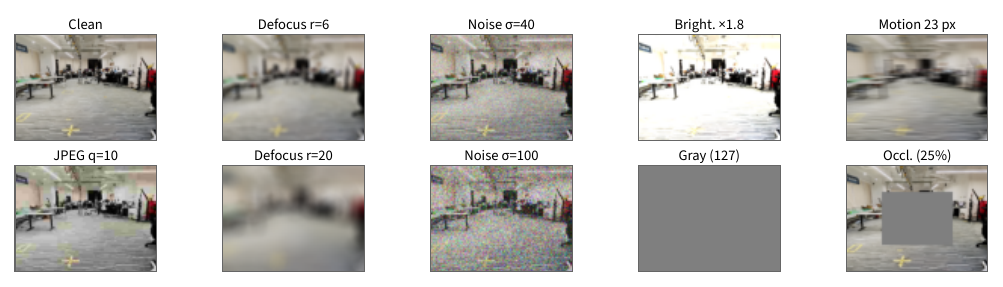}
\vspace{-5mm}
\caption{\textbf{All ten visual-intervention inputs on Route 01} ($240{\times}320$ left-eye RGB, applied immediately prior to policy inference while leaving proprioception, language, and controller inputs unchanged).}
\label{fig:supp_corruptions}
\vspace{-2mm}
\end{figure*}

\paragraph{Route-Aligned Coordinates and Corridor Excursion.}
For route $i$ with planar start $\mathbf{p}_i^{\mathrm{start}} \in \mathbb{R}^2$ and goal $\mathbf{p}_i^{\mathrm{goal}} \in \mathbb{R}^2$ ($d_i = \|\mathbf{p}_i^{\mathrm{goal}}-\mathbf{p}_i^{\mathrm{start}}\|_2 > 0$), let $\hat{\mathbf{e}}_i^{\parallel} = (\mathbf{p}_i^{\mathrm{goal}}-\mathbf{p}_i^{\mathrm{start}})/d_i = (e_{i,x}^{\parallel}, e_{i,y}^{\parallel})^\top$ and $\hat{\mathbf{e}}_i^{\perp} = (-e_{i,y}^{\parallel}, e_{i,x}^{\parallel})^\top$.
Each world-frame root position $\mathbf{p}_i(t)$ is transformed into along-track and lateral coordinates
\begin{equation}
\begin{aligned}
x_i(t) &= \bigl(\mathbf{p}_i(t)-\mathbf{p}_i^{\mathrm{start}}\bigr)^\top \hat{\mathbf{e}}_i^{\parallel}, \\
y_i(t) &= \bigl(\mathbf{p}_i(t)-\mathbf{p}_i^{\mathrm{start}}\bigr)^\top \hat{\mathbf{e}}_i^{\perp}.
\end{aligned}
\label{eq:supp_route_frame}
\end{equation}
Letting $t_i^{\star}$ denote first goal entry (or the final episode step if unsuccessful), the pre-arrival maximum lateral corridor excursion is
\begin{equation}
E_i = \max_{t \le t_i^{\star}} |y_i(t)|.
\label{eq:supp_excursion}
\end{equation}
Figure~\ref{fig:supp_route_atlas} displays all 15 routes under this canonical frame, and Table~\ref{tab:supp_task_decomp}c reports corridor departures ($E_i > 0.5$\,m): GR00T exceeds the 0.5\,m corridor on $4/15$ routes, the prior-initialized reference on $0/15$, and HumiAct on $1/15$, with $1/15$ fall each.
Notably, staying inside the lateral corridor alone is insufficient without sustained forward progress (the prior reference has $0/15$ departures yet reaches only $11/15$ goals vs.\ $13/15$ for HumiAct).

\paragraph{Exact Visual Corruption Specifications.}
\begin{table*}[t]
\centering
\caption{\textbf{Complete held-out UniAct representation analysis} (supporting Main Sec.~\mainref{sec:uniact_experiments}, Table~\mainref{tab:uniact}, and Fig.~\mainref{fig:representation_pca}). \textbf{(a)} 500 balanced held-out clips (100 per first action; source mix: AgiBotWorld 348 [69.6\%], Galaxea 127 [25.4\%], Humanoid 25 [5.0\%]) with 95\% bootstrap CIs (derangement min--max range over all 44 label permutations for the permutation control). \textbf{(b)} Per-action $H{=}1$ scores on the balanced panel. \textbf{(c)} Separate natural-distribution panel (489/500 manipulation-first clips; constant majority gives 97.8\% at $H{=}1$) and 4-class PCA projection statistics (seed 20260923).}
\label{tab:supp_uniact}
\vspace{-2mm}
\footnotesize
\setlength{\tabcolsep}{3pt}\renewcommand{\arraystretch}{0.94}
\begin{tabular*}{\textwidth}{@{\extracolsep{\fill}}lcccccc@{}}
\multicolumn{7}{@{}l}{\textbf{(a) Sequence discrimination across horizons (\%, balanced 500-clip held-out panel)}}\\[1pt]
\toprule
Method & $H{=}1$ Micro $\uparrow$ & $H{=}1$ Macro $\uparrow$ & $H{=}2$ Exact $\uparrow$ & $H{=}2$ Step $\uparrow$ & $H{=}3$ Exact $\uparrow$ & $H{=}3$ Step $\uparrow$ \\
\midrule
Random chance (analytic) & 20.0 & 20.0 & 4.0 & 20.0 & 0.8 & 20.0 \\
Train-marginal majority & 20.0 [16.6, 23.6] & 20.0 [20.0, 20.0] & 20.0 [16.6, 23.6] & 31.8 [28.4, 35.3] & 20.0 [16.6, 23.6] & 39.8 [36.4, 43.3] \\
Permuted action labels (44 derang.) & 3.0 [0.8, 5.0] & 3.0 [0.8, 5.0] & 0.7 [0.0, 1.6] & 4.6 [3.1, 6.0] & 0.3 [0.0, 0.6] & 5.2 [3.4, 6.7] \\
\textbf{UniAct (frozen predictor)} & \textbf{88.2 [85.4, 91.0]} & \textbf{88.2 [85.4, 91.0]} & \textbf{67.8 [63.6, 71.8]} & \textbf{81.5 [79.0, 84.0]} & \textbf{53.8 [49.6, 58.2]} & \textbf{79.3 [77.1, 81.6]} \\
\bottomrule
\end{tabular*}

\vspace{5pt}
\begin{minipage}[t]{0.50\textwidth}
\centering
\begin{tabular*}{\linewidth}{@{\extracolsep{\fill}}lrrrr@{}}
\multicolumn{5}{@{}l}{\textbf{(b) One-step per-action discrimination (balanced)}}\\[1pt]
\toprule
Behavior & Clips & Recall & Prec. & F1 \\
\midrule
Turn left & 100 & 90.0 & 93.8 & 91.8 \\
Move forward & 100 & 88.0 & 85.4 & 86.7 \\
Turn right & 100 & 88.0 & 86.3 & 87.1 \\
Move backward & 100 & 90.0 & 95.7 & 92.8 \\
Manipulate & 100 & 85.0 & 81.0 & 82.9 \\
\bottomrule
\end{tabular*}
\end{minipage}\hfill
\begin{minipage}[t]{0.47\textwidth}
\centering
\begin{tabular*}{\linewidth}{@{\extracolsep{\fill}}lr@{}}
\multicolumn{2}{@{}l}{\textbf{(c) Natural panel \& 4-class PCA geometry}}\\[1pt]
\toprule
Quantity & Value \\
\midrule
Natural $H{=}1$ Exact / Step & 92.0 [89.6, 94.4] / 92.0 [89.6, 94.4] \\
Natural $H{=}2$ Exact / Step & 75.8 [72.0, 79.4] / 87.6 [85.6, 89.5] \\
4-class argmax acc.\ (400 clips) & 89.75\% (359/400) \\
Mean true-action cosine & 0.6381 \\
PC1 / PC2 / Total var.\ (2D silh.) & 41.62\% / 34.62\% / 76.24\% (0.1731) \\
\bottomrule
\end{tabular*}
\end{minipage}
\end{table*}

Corruptions~\citep{hendrycks2019corruptions} are applied to the resized $240{\times}320$ left-eye RGB image immediately before policy inference (Fig.~\ref{fig:supp_corruptions}):
(i) \emph{Defocus blur}: convolution with a normalized disk kernel of radius $r{=}6$\,px (standard) or $r{=}20$\,px (strong);
(ii) \emph{Gaussian noise}: additive zero-mean noise with $\sigma{=}40$ (standard) or $\sigma{=}100$ (strong) on the 0--255 RGB scale, clipped to $[0,255]$ uint8;
(iii) \emph{Brightness}: multiplicative scaling by $1.8\times$ with clipping to 255;
(iv) \emph{Motion blur}: 23\,px horizontal box kernel;
(v) \emph{JPEG compression}: encode--decode round trip at quality $q{=}10$ with explicit RGB/BGR conversion;
(vi) \emph{Gray image}: all pixels replaced with constant RGB 127;
(vii) \emph{Central occlusion}: central half-width by half-height box ($25\%$ area) filled with RGB 127.

% ==========================================================================
\section{UniAct Predictive Prior and LAG Ablation Details}
\label{app:uniact_lag}
% ==========================================================================

\noindent\textbf{Claim anchor.}
\emph{Supports Main Sec.~\mainref{sec:latent_action_learning}--\mainref{sec:latent_action_guidance}, Sec.~\mainref{sec:uniact_experiments}, and Sec.~\mainref{sec:lag_ablation} (Tables~\mainref{tab:uniact} and \mainref{tab:lag_ablation}, Figs.~\mainref{fig:uniact_lag} and \mainref{fig:representation_pca}).}
\emph{Takeaway:} Action-conditioned predictive accuracy is balanced across all five behaviors (85--90\% one-step recall), and the LAG ablation confirms that downstream gains arise specifically from action-aligned future dynamics rather than generic feature regularization.

\begin{figure}[t]
\centering
\includegraphics[width=\linewidth]{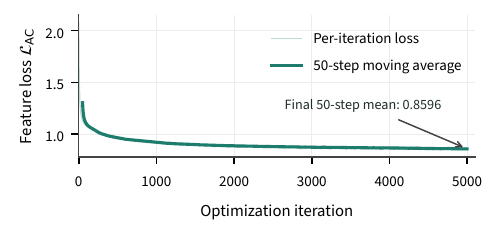}
\vspace{-5mm}
\caption{\textbf{Predictive-prior optimization curve} over 5,000 iterations (10 epochs $\times$ 500 steps; seed 20260916). Loss decreases smoothly from 2.0756 to a final 50-step mean of 0.8596.}
\label{fig:supp_prior_training}
\vspace{-2mm}
\end{figure}

\paragraph{Unified Action Mapping and Split Isolation.}
Every heterogeneous trajectory transition is mapped to one of five shared behaviors---turn left, move forward, turn right, move backward, and manipulate---encoded in indices 0--4 of a 7-D zero-padded one-hot condition $\mathbf{u}_t \in \{0,1\}^7$ (indices 5--6 remain zero).
Base translation and rotation identify navigation classes, while arm and hand motion identify manipulation; when both cues appear simultaneously, the largest normalized motion cue assigns the active class, and idle or ambiguous intervals are discarded so manipulation never acts as a catch-all.
Training, validation, and test partitions strictly separate source video, action-array, and source-parquet identifiers to prevent clip leakage.
Stage~1 trains $P_\psi$ over frozen ViT-giant features ($E_\phi$, 576 tokens of width 1408, 4 frames at 2\,Hz) with first-action sampling weights $[2,2,2,1,2]$ (left, forward, right, backward, manipulate) to counter manipulation dominance (Fig.~\ref{fig:supp_prior_training}).

\paragraph{Held-Out Sequence Discrimination and PCA Geometry.}
Table~\ref{tab:supp_uniact} provides complete confidence intervals, per-action scores, and natural-distribution controls for Main Table~\mainref{tab:uniact} and Fig.~\mainref{fig:representation_pca}.
For five equiprobable actions, random exact-sequence accuracy is $5^{-H}$ ($20.0\%, 4.0\%, 0.8\%$ for $H{=}1,2,3$) while per-step chance stays $20.0\%$.
On the balanced 500-clip panel (100 clips per first action), one-step recall spans $85.0$--$90.0\%$ across all five behaviors, verifying that minority navigation behaviors are represented as faithfully as manipulation.
At $H{=}3$, exact sequence accuracy reaches $53.8\%$ while per-step accuracy stays at $79.3\%$, showing that multi-step rollouts preserve action-specific dynamics even when a single step errs.
The separate natural-distribution panel (489/500 manipulation-first clips; Table~\ref{tab:supp_uniact}c) reaches $92.0\%$ at $H{=}1$ and $75.8\%$ exact at $H{=}2$, but is kept separate because a constant manipulation baseline already achieves $97.8\%$ first-action accuracy.

For the four-behavior geometry analysis (Main Fig.~\mainref{fig:representation_pca}a; 400 clips excluding backward motion), given clip context and observed future representation $\mathbf{Z}_i^+$ and candidate action prediction $\hat{\mathbf{Z}}_i^+(u)$, we compute candidate cosine scores and centered relative evidence:
\begin{equation}
s_{i,u}=\cos\!\bigl(\hat{\mathbf{Z}}_i^+(u),\mathbf{Z}_i^+\bigr),
\qquad
e_{i,u}=s_{i,u}-\frac{1}{4}\sum_{v} s_{i,v}.
\label{eq:supp_pca}
\end{equation}
Argmax over $s_{i,u}$ (computed \emph{before} PCA) selects the true behavior on $359/400$ clips ($89.75\%$, mean true-action cosine $0.6381$).
Standardizing each coordinate of $e_{i,\cdot}$ across the 400 clips and applying 2-component PCA (seed 20260923) explains $41.62\% + 34.62\% = 76.24\%$ of variance (2D silhouette $0.1731$).

\paragraph{LAG Distillation Interface and Ablation Design.}
In Stage~2, 32 auxiliary tokens (16 current, 16 future) are appended to the VLA backbone sequence, and student features $\mathbf{h}_t^{\mathrm{cur}}, \mathbf{h}_t^{\mathrm{fut}} \in \mathbb{R}^{16\times 2048}$ are extracted from the third-from-last language-backbone layer.
A shared linear projector $G:\mathbb{R}^{2048}\!\to\!\mathbb{R}^{1408}$ maps them to the teacher width, while learned attention poolers $Q_{\mathrm{cur}}, Q_{\mathrm{fut}}$ compress the 576 teacher patch tokens into 16 target tokens, trained with equal unit weights ($\lambda_{\mathrm{cur}}{=}\lambda_{\mathrm{fut}}{=}1.0$, matching Main Eq.~\mainref{eq:latent_action_guidance}).
All ablation variants in Main Table~\mainref{tab:lag_ablation} undergo the exact same Stage~3 fine-tuning recipe:
\begin{itemize}
\item \emph{Current + Aligned} sets $\lambda_{\mathrm{fut}}{=}0$, keeping only current-frame teacher features; it matches HumiAct on Terra-guided navigation ($53.3\%$) but trails on direct navigation ($73.3\%$ vs.\ $86.7\%$) and five-corruption mean ($78.7\%$ vs.\ $84.0\%$).
\item \emph{Future + None} and \emph{Current + future + None} remove the UniAct action condition $\mathbf{u}_t$ from $P_\psi$, testing unconditioned predictive distillation; both underperform HumiAct ($66.7\%/73.3\%$ direct; $76.0\%/80.0\%$ corruption mean).
\item \emph{Current + future + Shuffled} permutes teacher action labels during distillation while keeping architecture and budget identical; performance drops sharply below even unguided GR00T ($46.7\%$ vs.\ $66.7\%$ direct; $74.6\%$ vs.\ $77.3\%$ corruption mean), confirming that mis-aligned predictive targets induce destructive interference and that HumiAct's gains specifically require action-aligned dynamics.
\end{itemize}

% ==========================================================================
\section{Task Planner, Physical-Robot Evaluation, and Qualitative Rollouts}
\label{app:planner_real}
% ==========================================================================

\noindent\textbf{Claim anchor.}
\emph{Supports Main Sec.~\mainref{sec:unified_control}, Sec.~\mainref{sec:data_to_robot}, and Sec.~\mainref{sec:policy_planner}--\mainref{sec:conclusion} (Table~\mainref{tab:planner_main}, Figs.~\mainref{fig:intro_overview} and~\mainref{fig:supp_navigation_frames}).}
\emph{Takeaway:} Varying only the high-level planner reveals a strong SR--exposure trade-off (6.7\% to 93.3\% SR with the exact same motor policy), while physical-robot trials and ten representative rollouts substantiate closed-loop walking, bar-counter approach, grasping/delivery, and whole-body chair pushing.

\paragraph{Closed-Loop Planner Contract.}
After each executed action chunk, the planner receives the full episode observation history and outputs a structured JSON decision containing \texttt{action\_id}, \texttt{objective}, \texttt{remaining\_plan}, \texttt{decision} $\in \{\text{retain}, \text{advance}, \text{revise}\}$, \texttt{reason}, and \texttt{visual\_state}.
All six planners in Main Table~\mainref{tab:planner_main} drive the exact same HumiAct checkpoint and whole-body controller on the same 15 obstacle routes; GPT-6 Astra uses low reasoning effort, GPT-6.1 Sol uses high reasoning effort, and the remaining four planners use default settings.
As shown in Fig.~\ref{fig:supp_planner_funnel}a, stronger visual-spatial planners (GPT-6 Astra: $14/15$, SPL $0.716$; GPT-6.1 Sol: $11/15$, SPL $0.569$) unlock substantially higher obstacle-navigation performance from the fixed motor policy than the default GPT-5.6 Terra planner ($8/15$, SPL $0.410$), confirming that Main Table~\mainref{tab:main_results} is conservative with respect to planner quality.

\begin{table}[t]
\centering
\caption{\textbf{Physical-robot primitive motion trials} (supporting Main Sec.~\mainref{sec:policy_planner}). Planner disabled; 1 denotes successful execution and 0 denotes failure across three independent repetitions per command.}
\label{tab:supp_real_motion}
\footnotesize
\setlength{\tabcolsep}{2.8pt}
\begin{tabular*}{\linewidth}{@{\extracolsep{\fill}}l cccc cccc@{}}
\toprule
& \multicolumn{4}{c}{GR00T} & \multicolumn{4}{c}{Nav.-adapted policy} \\
\cmidrule(lr){2-5}\cmidrule(lr){6-9}
Command & T1 & T2 & T3 & Succ. & T1 & T2 & T3 & Succ. \\
\midrule
Turn left & 1 & 1 & 1 & 3/3 & 1 & 1 & 1 & 3/3 \\
Turn right & 1 & 1 & 1 & 3/3 & 1 & 1 & 1 & 3/3 \\
Move forward & 0 & 1 & 0 & 1/3 & 1 & 1 & 0 & 2/3 \\
\bottomrule
\end{tabular*}
\end{table}

\begin{figure*}[t]
\centering
\includegraphics[width=0.86\textwidth]{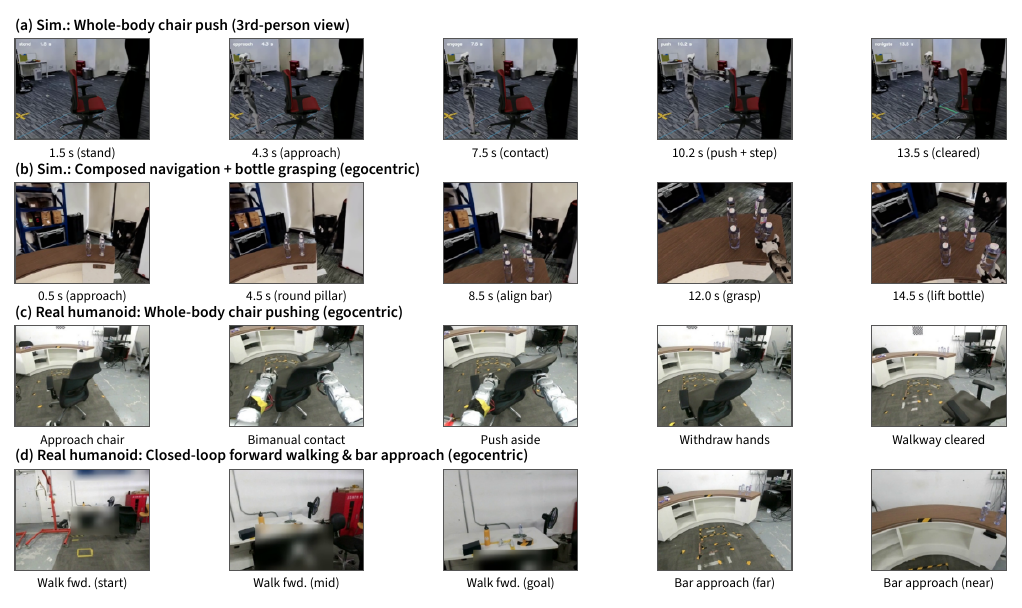}
\vspace{-4mm}
\caption{\textbf{Qualitative filmstrips across simulation and real-humanoid deployment.} \textbf{(a)} Simulated whole-body chair pushing (3rd-person): walking approach, bimanual contact, stepping while pushing, and clearing the walkway. \textbf{(b)} Simulated composed navigation and bottle grasping (32-trial benchmark). \textbf{(c)} Real 33-DoF humanoid bimanual chair pushing in front of the bar counter (egocentric). \textbf{(d)} Real 33-DoF humanoid closed-loop forward walking (Table~\ref{tab:supp_real_motion}, Trial~2) and bar-counter approach (egocentric).}
\label{fig:supp_filmstrips}
\end{figure*}
\begin{table*}[t]
\centering
\caption{\textbf{Summary of qualitative simulation and physical-humanoid rollouts.} Representative closed-loop demonstrations across simulation and real-robot evaluation corresponding to the main text and Fig.~\ref{fig:supp_filmstrips}.}
\label{tab:supp_media}
\vspace{-2mm}
\scriptsize
\setlength{\tabcolsep}{2.5pt}\renewcommand{\arraystretch}{0.92}
\begin{tabular*}{\textwidth}{@{\extracolsep{\fill}}>{\raggedright\arraybackslash}p{.36\textwidth}p{.055\textwidth}p{.095\textwidth}p{.46\textwidth}@{}}
\toprule
Rollout scenario & Dom. & Anchor & Content summary \\
\midrule
1.\ Direct-navigation comparison (Route 04) & Sim & Sec.~\mainref{sec:whole_body_results} & GR00T vs.\ HumiAct on direct-nav.\ Route 04 (Fig.~\mainref{fig:supp_navigation_frames}a). \\
2.\ Whole-body chair push (dual-view) & Sim & Fig.~\mainref{fig:intro_overview}b & Dual-view chair push: approach, bimanual push, traversal. \\
3.\ Chair push and bar approach (egocentric) & Sim & Sec.~\mainref{sec:unified_control} & Egocentric bimanual chair push + continuous bar approach. \\
4.\ Composed navigation and grasp (Trial~1) & Sim & Table~\mainref{tab:main_results} & Composed 32-trial Trial~1: pillar approach + bottle grasp. \\
5.\ Bimanual chair push (cleanroom) & Sim & Sec.~\mainref{sec:data_to_robot} & Cleanroom physics check of bimanual push and arm swing. \\
6.\ Physical-humanoid chair pushing & Real & Sec.~\mainref{sec:conclusion} & Physical 33-DoF humanoid: two chair-push trials. \\
7.\ Physical-humanoid walking and bar approach & Real & Sec.~\mainref{sec:policy_planner} & Physical humanoid: forward walk (Table~\ref{tab:supp_real_motion}, T2) + bar approach. \\
8.\ Composed navigation and grasp (Trial~2) & Sim & Table~\mainref{tab:main_results} & Composed 32-trial Trial~2: pillar turn, alignment, bottle lift. \\
9.\ Closed-loop banana grasping & Sim & Sec.~\mainref{sec:data_to_robot} & Closed-loop banana top-pinch: alignment and 28\,cm lift. \\
10.\ Closed-loop milk-carton delivery & Sim & Sec.~\mainref{sec:data_to_robot} & Full closed-loop milk-carton delivery (0 collisions, complete). \\
\bottomrule
\end{tabular*}
\end{table*}

\paragraph{Physical-Robot Execution Stack and Primitive Trials.}
On the physical 33-DoF humanoid (37-D proprioceptive state, synchronized head-mounted cameras), the shared VLA runs on an NVIDIA RTX 5000 Ada at 10\,Hz (BF16, 4 flow-integration steps, 30-step chunks) and communicates with the onboard ROS~2 client via ZeroMQ.
Buffered references are aligned to the execution clock, smoothly blended across chunk transitions, and streamed at 50\,Hz to the whole-body tracker with synchronized head and hand commands.
Table~\ref{tab:supp_real_motion} records all per-trial outcomes for the planner-free motion-command evaluation referenced in Main Sec.~\mainref{sec:policy_planner}.
Both GR00T and the navigation-adapted policy complete $3/3$ left turns and $3/3$ right turns; on forward walking, the adapted policy succeeds in $2/3$ trials vs.\ $1/3$ for GR00T (the single adapted-policy failure resulted from an initial heading drift before gait stabilization).
As emphasized in Sec.~\ref{app:protocol}, these three-repetition primitive trials provide controlled hardware verification of motion execution and are distinct from full obstacle or composed navigation trials.

\paragraph{Qualitative Rollout Summary.}
Figure~\ref{fig:supp_filmstrips} presents keyframe sequences from both physics-based simulation and physical-robot rollouts, and Table~\ref{tab:supp_media} summarizes the ten representative qualitative demonstrations across simulation and real-humanoid evaluation (supporting Main Figs.~\mainref{fig:intro_overview} and~\mainref{fig:supp_navigation_frames}, and Sec.~\mainref{sec:whole_body_results}--\mainref{sec:conclusion}).


\begin{thebibliography}{92}
\providecommand{\natexlab}[1]{#1}
\providecommand{\url}[1]{\texttt{#1}}
\expandafter\ifx\csname urlstyle\endcsname\relax
  \providecommand{\doi}[1]{doi: #1}\else
  \providecommand{\doi}{doi: \begingroup \urlstyle{rm}\Url}\fi

\bibitem[{AgiBot World Contributors} et~al.(2025){AgiBot World Contributors}, Bu, Cai, Chen, Cui, Ding, Feng, Gao, He, Hu, Huang, Jiang, Jiang, Jing, Li, Li, Liu, Liu, Lu, Luo, Luo, Mu, Niu, Pan, Pang, Qiao, Ren, Ruan, Shan, Shen, Shi, Shi, Shi, Sima, Song, Wang, Wang, Wei, Xie, Xu, Yan, Yang, Yang, Yang, Yao, Zeng, Zhang, Zhang, Zhao, Zhao, Zhao, and Zhu]{bu2025agibot}
{AgiBot World Contributors}, Qingwen Bu, Jisong Cai, Li Chen, Xiuqi Cui, Yan Ding, Siyuan Feng, Shenyuan Gao, Xindong He, Xuan Hu, Xu Huang, Shu Jiang, Yuxin Jiang, Cheng Jing, Hongyang Li, Jialu Li, Chiming Liu, Yi Liu, Yuxiang Lu, Jianlan Luo, Ping Luo, Yao Mu, Yuehan Niu, Yixuan Pan, Jiangmiao Pang, Yu Qiao, Guanghui Ren, Cheng Ruan, Jiaqi Shan, Yongjian Shen, Chengshi Shi, Mingkang Shi, Modi Shi, Chonghao Sima, Jianheng Song, Huijie Wang, Wenhao Wang, Dafeng Wei, Chengen Xie, Guo Xu, Junchi Yan, Cunbiao Yang, Lei Yang, Shukai Yang, Maoqing Yao, Jia Zeng, Chi Zhang, Qinglin Zhang, Bin Zhao, Chengyue Zhao, Jiaqi Zhao, and Jianchao Zhu.
\newblock {AgiBot World Colosseo: A Large-scale Manipulation Platform for Scalable and Intelligent Embodied Systems}.
\newblock arXiv preprint arXiv:2503.06669, 2025.

\bibitem[Ahn et~al.(2022)Ahn, Brohan, Brown, Chebotar, Cortes, David, Finn, Fu, Gopalakrishnan, Hausman, Herzog, Ho, Hsu, Ibarz, Ichter, Irpan, Jang, Ruano, Jeffrey, Jesmonth, Joshi, Julian, Kalashnikov, Kuang, Lee, Levine, Lu, Luu, Parada, Pastor, Quiambao, Rao, Rettinghouse, Reyes, Sermanet, Sievers, Tan, Toshev, Vanhoucke, Xia, Xiao, Xu, Xu, Yan, and Zeng]{ahn2022saycan}
Michael Ahn, Anthony Brohan, Noah Brown, Yevgen Chebotar, Omar Cortes, Byron David, Chelsea Finn, Chuyuan Fu, Keerthana Gopalakrishnan, Karol Hausman, Alex Herzog, Daniel Ho, Jasmine Hsu, Julian Ibarz, Brian Ichter, Alex Irpan, Eric Jang, Rosario~Jauregui Ruano, Kyle Jeffrey, Sally Jesmonth, Nikhil~J Joshi, Ryan Julian, Dmitry Kalashnikov, Yuheng Kuang, Kuang-Huei Lee, Sergey Levine, Yao Lu, Linda Luu, Carolina Parada, Peter Pastor, Jornell Quiambao, Kanishka Rao, Jarek Rettinghouse, Diego Reyes, Pierre Sermanet, Nicolas Sievers, Clayton Tan, Alexander Toshev, Vincent Vanhoucke, Fei Xia, Ted Xiao, Peng Xu, Sichun Xu, Mengyuan Yan, and Andy Zeng.
\newblock {Do As I Can, Not As I Say: Grounding Language in Robotic Affordances}.
\newblock arXiv preprint arXiv:2204.01691, 2022.

\bibitem[Anderson et~al.(2018{\natexlab{a}})Anderson, Chang, Chaplot, Dosovitskiy, Gupta, Koltun, Kosecka, Malik, Mottaghi, Savva, and Zamir]{anderson2018evaluation}
Peter Anderson, Angel~X. Chang, Devendra~Singh Chaplot, Alexey Dosovitskiy, Saurabh Gupta, Vladlen Koltun, Jana Kosecka, Jitendra Malik, Roozbeh Mottaghi, Manolis Savva, and Amir~R. Zamir.
\newblock {On Evaluation of Embodied Navigation Agents}.
\newblock arXiv preprint arXiv:1807.06757, 2018{\natexlab{a}}.

\bibitem[Anderson et~al.(2018{\natexlab{b}})Anderson, Wu, Teney, Bruce, Johnson, S{\"u}nderhauf, Reid, Gould, and van~den Hengel]{anderson2018vln}
Peter Anderson, Qi Wu, Damien Teney, Jake Bruce, Mark Johnson, Niko S{\"u}nderhauf, Ian Reid, Stephen Gould, and Anton van~den Hengel.
\newblock {Vision-and-Language Navigation: Interpreting visually-grounded navigation instructions in real environments}.
\newblock In \emph{Proceedings of the IEEE Conference on Computer Vision and Pattern Recognition}, pages 3674--3683, 2018{\natexlab{b}}.

\bibitem[Assran et~al.(2023)Assran, Duval, Misra, Bojanowski, Vincent, Rabbat, LeCun, and Ballas]{assran2023ijepa}
Mahmoud Assran, Quentin Duval, Ishan Misra, Piotr Bojanowski, Pascal Vincent, Michael Rabbat, Yann LeCun, and Nicolas Ballas.
\newblock {Self-Supervised Learning from Images with a Joint-Embedding Predictive Architecture}.
\newblock In \emph{IEEE/CVF Conference on Computer Vision and Pattern Recognition}, pages 15619--15629, 2023.

\bibitem[Assran et~al.(2025)Assran, Bardes, Fan, Garrido, Howes, Komeili, Muckley, Rizvi, Roberts, Sinha, Zholus, Arnaud, Gejji, Martin, Hogan, Dugas, Bojanowski, Khalidov, Labatut, Massa, Szafraniec, Krishnakumar, Li, Ma, Chandar, Meier, LeCun, Rabbat, and Ballas]{assran2025vjepa2}
Mido Assran, Adrien Bardes, David Fan, Quentin Garrido, Russell Howes, Mojtaba Komeili, Matthew~J. Muckley, Ammar Rizvi, Claire Roberts, Koustuv Sinha, Artem Zholus, Sergio Arnaud, Abha Gejji, Ada Martin, Francois~Robert Hogan, Daniel Dugas, Piotr Bojanowski, Vasil Khalidov, Patrick Labatut, Francisco Massa, Marc Szafraniec, Kapil Krishnakumar, Yong Li, Xiaodong Ma, Sarath Chandar, Franziska Meier, Yann LeCun, Michael Rabbat, and Nicolas Ballas.
\newblock {V-JEPA 2: Self-Supervised Video Models Enable Understanding, Prediction and Planning}.
\newblock arXiv preprint arXiv:2506.09985, 2025.

\bibitem[Bardes et~al.(2024)Bardes, Garrido, Ponce, Chen, Rabbat, LeCun, Assran, and Ballas]{bardes2024vjepa}
Adrien Bardes, Quentin Garrido, Jean Ponce, Xinlei Chen, Michael Rabbat, Yann LeCun, Mido Assran, and Nicolas Ballas.
\newblock {Revisiting Feature Prediction for Learning Visual Representations from Video}.
\newblock \emph{Transactions on Machine Learning Research}, 2024.

\bibitem[Black et~al.(2024)Black, Brown, Driess, et~al.]{black2024pi0}
Kevin Black, Noah Brown, Danny Driess, et~al.
\newblock {$\pi_0$: A Vision-Language-Action Flow Model for General Robot Control}.
\newblock arXiv preprint arXiv:2410.24164, 2024.

\bibitem[Brohan et~al.(2023)Brohan, Brown, Carbajal, et~al.]{brohan2023rt1}
Anthony Brohan, Noah Brown, Justice Carbajal, et~al.
\newblock {RT-1: Robotics Transformer for Real-World Control at Scale}.
\newblock In \emph{Robotics: Science and Systems}, 2023.

\bibitem[Bruce et~al.(2024)Bruce, Dennis, Edwards, Parker-Holder, Shi, Hughes, Lai, Mavalankar, Steigerwald, Apps, Aytar, Bechtle, Behbahani, Chan, Heess, Gonzalez, Osindero, Ozair, Reed, Zhang, Zolna, Clune, de~Freitas, Singh, and Rockt{\"a}schel]{bruce2024genie}
Jake Bruce, Michael~D. Dennis, Ashley Edwards, Jack Parker-Holder, Yuge Shi, Edward Hughes, Matthew Lai, Aditi Mavalankar, Richie Steigerwald, Chris Apps, Yusuf Aytar, Sarah Maria~Elisabeth Bechtle, Feryal Behbahani, Stephanie C.~Y. Chan, Nicolas Heess, Lucy Gonzalez, Simon Osindero, Sherjil Ozair, Scott Reed, Jingwei Zhang, Konrad Zolna, Jeff Clune, Nando de Freitas, Satinder Singh, and Tim Rockt{\"a}schel.
\newblock {Genie: Generative Interactive Environments}.
\newblock In \emph{Proceedings of the 41st International Conference on Machine Learning}, pages 4603--4623, 2024.

\bibitem[Caron et~al.(2021)Caron, Touvron, Misra, J\'{e}gou, Mairal, Bojanowski, and Joulin]{caron2021dino}
Mathilde Caron, Hugo Touvron, Ishan Misra, Herv\'{e} J\'{e}gou, Julien Mairal, Piotr Bojanowski, and Armand Joulin.
\newblock {Emerging Properties in Self-Supervised Vision Transformers}.
\newblock In \emph{IEEE/CVF International Conference on Computer Vision}, pages 9630--9640, 2021.

\bibitem[Chen et~al.(2021)Chen, Guhur, Schmid, and Laptev]{chen2021hamt}
Shizhe Chen, Pierre-Louis Guhur, Cordelia Schmid, and Ivan Laptev.
\newblock {History Aware Multimodal Transformer for Vision-and-Language Navigation}.
\newblock In \emph{Advances in Neural Information Processing Systems}, pages 5834--5847, 2021.

\bibitem[Chi et~al.(2023)Chi, Feng, Du, Xu, Cousineau, Burchfiel, and Song]{chi2023diffusion}
Cheng Chi, Siyuan Feng, Yilun Du, Zhenjia Xu, Eric Cousineau, Benjamin Burchfiel, and Shuran Song.
\newblock {Diffusion Policy: Visuomotor Policy Learning via Action Diffusion}.
\newblock In \emph{Robotics: Science and Systems}, 2023.

\bibitem[Cobbe et~al.(2019)Cobbe, Klimov, Hesse, Kim, and Schulman]{cobbe2019generalization}
Karl Cobbe, Oleg Klimov, Chris Hesse, Taehoon Kim, and John Schulman.
\newblock {Quantifying Generalization in Reinforcement Learning}.
\newblock In \emph{International Conference on Machine Learning}, pages 1282--1289, 2019.

\bibitem[Dasari et~al.(2020)Dasari, Ebert, Tian, Nair, Bucher, Schmeckpeper, Singh, Levine, and Finn]{dasari2019robonet}
Sudeep Dasari, Frederik Ebert, Stephen Tian, Suraj Nair, Bernadette Bucher, Karl Schmeckpeper, Siddharth Singh, Sergey Levine, and Chelsea Finn.
\newblock {RoboNet: Large-Scale Multi-Robot Learning}.
\newblock In \emph{Proceedings of the Conference on Robot Learning}, pages 885--897. PMLR, 2020.

\bibitem[Driess et~al.(2023)Driess, Xia, Sajjadi, Lynch, Chowdhery, Ichter, Wahid, Tompson, Vuong, Yu, Huang, Chebotar, Sermanet, Duckworth, Levine, Vanhoucke, Hausman, Toussaint, Greff, Zeng, Mordatch, and Florence]{driess2023palme}
Danny Driess, Fei Xia, Mehdi S.~M. Sajjadi, Corey Lynch, Aakanksha Chowdhery, Brian Ichter, Ayzaan Wahid, Jonathan Tompson, Quan Vuong, Tianhe Yu, Wenlong Huang, Yevgen Chebotar, Pierre Sermanet, Daniel Duckworth, Sergey Levine, Vincent Vanhoucke, Karol Hausman, Marc Toussaint, Klaus Greff, Andy Zeng, Igor Mordatch, and Pete Florence.
\newblock {PaLM-E: An Embodied Multimodal Language Model}.
\newblock In \emph{International Conference on Machine Learning}, pages 8469--8488, 2023.

\bibitem[Efron and Tibshirani(1993)]{efron1993bootstrap}
Bradley Efron and Robert~J. Tibshirani.
\newblock \emph{{An Introduction to the Bootstrap}}.
\newblock Chapman \& Hall/CRC, 1993.

\bibitem[Fisher(1922)]{fisher1922interpretation}
R.~A. Fisher.
\newblock {On the Interpretation of $\chi^2$ from Contingency Tables, and the Calculation of $P$}.
\newblock \emph{Journal of the Royal Statistical Society}, 85\penalty0 (1):\penalty0 87--94, 1922.

\bibitem[Fried et~al.(2018)Fried, Hu, Cirik, Rohrbach, Andreas, Morency, Berg-Kirkpatrick, Saenko, Klein, and Darrell]{fried2018speaker}
Daniel Fried, Ronghang Hu, Volkan Cirik, Anna Rohrbach, Jacob Andreas, Louis-Philippe Morency, Taylor Berg-Kirkpatrick, Kate Saenko, Dan Klein, and Trevor Darrell.
\newblock {Speaker-Follower Models for Vision-and-Language Navigation}.
\newblock In \emph{Advances in Neural Information Processing Systems}, pages 3318--3329, 2018.

\bibitem[Fu et~al.(2024{\natexlab{a}})Fu, Zhao, Wu, Wetzstein, and Finn]{fu2024humanplus}
Zipeng Fu, Qingqing Zhao, Qi Wu, Gordon Wetzstein, and Chelsea Finn.
\newblock {HumanPlus: Humanoid Shadowing and Imitation from Humans}.
\newblock In \emph{Conference on Robot Learning}, pages 2828--2844, 2024{\natexlab{a}}.

\bibitem[Fu et~al.(2024{\natexlab{b}})Fu, Zhao, and Finn]{fu2024mobilealoha}
Zipeng Fu, Tony~Z. Zhao, and Chelsea Finn.
\newblock {Mobile ALOHA: Learning Bimanual Mobile Manipulation with Low-Cost Whole-Body Teleoperation}.
\newblock In \emph{Conference on Robot Learning}, pages 4066--4083, 2024{\natexlab{b}}.

\bibitem[Ghosh et~al.(2024)Ghosh, Walke, Pertsch, Black, Mees, Dasari, Hejna, Kreiman, Xu, Luo, Tan, Chen, Vuong, Xiao, Sanketi, Sadigh, Finn, and Levine]{ghosh2024octo}
Dibya Ghosh, Homer~Rich Walke, Karl Pertsch, Kevin Black, Oier Mees, Sudeep Dasari, Joey Hejna, Tobias Kreiman, Charles Xu, Jianlan Luo, You~Liang Tan, Lawrence~Yunliang Chen, Quan Vuong, Ted Xiao, Pannag~R. Sanketi, Dorsa Sadigh, Chelsea Finn, and Sergey Levine.
\newblock {Octo: An Open-Source Generalist Robot Policy}.
\newblock In \emph{Robotics: Science and Systems}, 2024.

\bibitem[Gu et~al.(2024)Gu, Wang, and Chen]{gu2024humanoidgym}
Xinyang Gu, Yen-Jen Wang, and Jianyu Chen.
\newblock {Humanoid-Gym: Reinforcement Learning for Humanoid Robot with Zero-Shot Sim2Real Transfer}.
\newblock arXiv preprint arXiv:2404.05695, 2024.

\bibitem[Ha and Schmidhuber(2018)]{ha2018worldmodels}
David Ha and J\"{u}rgen Schmidhuber.
\newblock {World Models}.
\newblock arXiv preprint arXiv:1803.10122, 2018.

\bibitem[Hafner et~al.(2019)Hafner, Lillicrap, Fischer, Villegas, Ha, Lee, and Davidson]{hafner2019planet}
Danijar Hafner, Timothy Lillicrap, Ian Fischer, Ruben Villegas, David Ha, Honglak Lee, and James Davidson.
\newblock {Learning Latent Dynamics for Planning from Pixels}.
\newblock In \emph{International Conference on Machine Learning}, pages 2555--2565, 2019.

\bibitem[Hafner et~al.(2020)Hafner, Lillicrap, Ba, and Norouzi]{hafner2020dreamer}
Danijar Hafner, Timothy Lillicrap, Jimmy Ba, and Mohammad Norouzi.
\newblock {Dream to Control: Learning Behaviors by Latent Imagination}.
\newblock In \emph{International Conference on Learning Representations}, 2020.

\bibitem[Hafner et~al.(2021)Hafner, Lillicrap, Norouzi, and Ba]{hafner2021dreamerv2}
Danijar Hafner, Timothy Lillicrap, Mohammad Norouzi, and Jimmy Ba.
\newblock {Mastering Atari with Discrete World Models}.
\newblock In \emph{International Conference on Learning Representations}, 2021.

\bibitem[Han et~al.(2026)Han, Hao, Ma, Zhumakhanova, Radionova, Zhang, Chang, Liang, and Laptev]{han2026implicit}
Mingfei Han, Haihong Hao, Liang Ma, Kamila Zhumakhanova, Ekaterina Radionova, Jingyi Zhang, Xiaojun Chang, Xiaodan Liang, and Ivan Laptev.
\newblock {Implicit Geometry Representations for Vision-and-Language Navigation from Web Videos}.
\newblock arXiv preprint arXiv:2603.09259, 2026.

\bibitem[Hao et~al.(2026{\natexlab{a}})Hao, Chen, Han, Li, An, Yang, Li, and Chang]{hao2026latentpilot}
Haihong Hao, Lei Chen, Mingfei Han, Changlin Li, Dong An, Yuqiang Yang, Zhihui Li, and Xiaojun Chang.
\newblock {LatentPilot: Scene-Aware Vision-and-Language Navigation by Dreaming Ahead with Latent Visual Reasoning}.
\newblock In \emph{European Conference on Computer Vision}, pages 75--93, 2026{\natexlab{a}}.

\bibitem[Hao et~al.(2026{\natexlab{b}})Hao, Chen, Han, Yan, An, Yang, Li, Guo, Khan, and Chang]{hao2026seer}
Haihong Hao, Lei Chen, Mingfei Han, Chenglong Yan, Dong An, Yuqiang Yang, Changlin Li, Minghao Guo, Salman Khan, and Xiaojun Chang.
\newblock {Seeing Together, Acting Apart: Shared Environmental Understanding for Multi-Robot Navigation}.
\newblock In \emph{Advances in Neural Information Processing Systems}, 2026{\natexlab{b}}.

\bibitem[Hao et~al.(2026{\natexlab{c}})Hao, Han, Li, Li, and Chang]{hao2026conav}
Haihong Hao, Mingfei Han, Changlin Li, Zhihui Li, and Xiaojun Chang.
\newblock {CoNav: Collaborative Cross-Modal Reasoning for Embodied Navigation}.
\newblock \emph{IEEE Transactions on Pattern Analysis and Machine Intelligence}, 48\penalty0 (11):\penalty0 14239--14253, 2026{\natexlab{c}}.

\bibitem[He et~al.(2022)He, Chen, Xie, Li, Doll\'{a}r, and Girshick]{he2022mae}
Kaiming He, Xinlei Chen, Saining Xie, Yanghao Li, Piotr Doll\'{a}r, and Ross Girshick.
\newblock {Masked Autoencoders Are Scalable Vision Learners}.
\newblock In \emph{IEEE/CVF Conference on Computer Vision and Pattern Recognition}, pages 15979--15988, 2022.

\bibitem[He et~al.(2024{\natexlab{a}})He, Luo, He, Xiao, Zhang, Zhang, Kitani, Liu, and Shi]{he2024omnih2o}
Tairan He, Zhengyi Luo, Xialin He, Wenli Xiao, Chong Zhang, Weinan Zhang, Kris Kitani, Changliu Liu, and Guanya Shi.
\newblock {OmniH2O: Universal and Dexterous Human-to-Humanoid Whole-Body Teleoperation and Learning}.
\newblock In \emph{Conference on Robot Learning}, pages 1516--1540, 2024{\natexlab{a}}.

\bibitem[He et~al.(2024{\natexlab{b}})He, Luo, Xiao, Zhang, Kitani, Liu, and Shi]{he2024h2o}
Tairan He, Zhengyi Luo, Wenli Xiao, Chong Zhang, Kris Kitani, Changliu Liu, and Guanya Shi.
\newblock {Learning Human-to-Humanoid Real-Time Whole-Body Teleoperation}.
\newblock arXiv preprint arXiv:2403.04436, 2024{\natexlab{b}}.

\bibitem[He et~al.(2025)He, Xiao, Lin, Luo, Xu, Jiang, Kautz, Liu, Shi, Wang, Fan, and Zhu]{he2025hover}
Tairan He, Wenli Xiao, Toru Lin, Zhengyi Luo, Zhenjia Xu, Zhenyu Jiang, Jan Kautz, Changliu Liu, Guanya Shi, Xiaolong Wang, Linxi Fan, and Yuke Zhu.
\newblock {HOVER: Versatile Neural Whole-Body Controller for Humanoid Robots}.
\newblock In \emph{IEEE International Conference on Robotics and Automation}, pages 9989--9996, 2025.

\bibitem[Hendrycks and Dietterich(2019)]{hendrycks2019corruptions}
Dan Hendrycks and Thomas Dietterich.
\newblock {Benchmarking Neural Network Robustness to Common Corruptions and Perturbations}.
\newblock In \emph{International Conference on Learning Representations}, 2019.

\bibitem[Hinton et~al.(2015)Hinton, Vinyals, and Dean]{hinton2015distill}
Geoffrey Hinton, Oriol Vinyals, and Jeff Dean.
\newblock {Distilling the Knowledge in a Neural Network}.
\newblock arXiv preprint arXiv:1503.02531, 2015.

\bibitem[Hong et~al.(2021)Hong, Wu, Qi, Opazo, and Gould]{hong2021vlnbert}
Yicong Hong, Qi Wu, Yuankai Qi, Cristian~Rodriguez Opazo, and Stephen Gould.
\newblock {VLN BERT: A Recurrent Vision-and-Language BERT for Navigation}.
\newblock In \emph{IEEE/CVF Conference on Computer Vision and Pattern Recognition}, pages 1643--1653, 2021.

\bibitem[Huang et~al.(2022)Huang, Xia, Xiao, Chan, Liang, Florence, Zeng, Tompson, Mordatch, Chebotar, Sermanet, Brown, Jackson, Luu, Levine, Hausman, and Ichter]{huang2022inner}
Wenlong Huang, Fei Xia, Ted Xiao, Harris Chan, Jacky Liang, Pete Florence, Andy Zeng, Jonathan Tompson, Igor Mordatch, Yevgen Chebotar, Pierre Sermanet, Noah Brown, Tomas Jackson, Linda Luu, Sergey Levine, Karol Hausman, and Brian Ichter.
\newblock {Inner Monologue: Embodied Reasoning through Planning with Language Models}.
\newblock In \emph{Conference on Robot Learning}, 2022.

\bibitem[Huang et~al.(2023)Huang, Wang, Zhang, Li, Wu, and Fei-Fei]{huang2023voxposer}
Wenlong Huang, Chen Wang, Ruohan Zhang, Yunzhu Li, Jiajun Wu, and Li Fei-Fei.
\newblock {VoxPoser: Composable 3D Value Maps for Robotic Manipulation with Language Models}.
\newblock In \emph{Conference on Robot Learning}, pages 540--562, 2023.

\bibitem[Hwangbo et~al.(2019)Hwangbo, Lee, Dosovitskiy, Bellicoso, Tsounis, Koltun, and Hutter]{hwangbo2019agile}
Jemin Hwangbo, Joonho Lee, Alexey Dosovitskiy, Dario Bellicoso, Vassilios Tsounis, Vladlen Koltun, and Marco Hutter.
\newblock {Learning agile and dynamic motor skills for legged robots}.
\newblock \emph{Science Robotics}, 4\penalty0 (26):\penalty0 eaau5872, 2019.

\bibitem[Jeong et~al.(2026)Jeong, Chun, and Kim]{jeong2026vila}
Youngjoon Jeong, Junha Chun, and Taesup Kim.
\newblock {Learning to Act Robustly with View-Invariant Latent Actions}.
\newblock In \emph{Proceedings of the IEEE/CVF Conference on Computer Vision and Pattern Recognition}, pages 6781--6790, 2026.

\bibitem[Jiang et~al.(2026)Jiang, Chen, Bu, Chen, Shi, Zhang, Li, Suo, Wang, Peng, and Li]{jiang2025wholebodyvla}
Haoran Jiang, Jin Chen, Qingwen Bu, Li Chen, Modi Shi, Yanjie Zhang, Delong Li, Chuanzhe Suo, Chuang Wang, Zhihui Peng, and Hongyang Li.
\newblock {WholeBodyVLA: Towards Unified Latent VLA for Whole-Body Loco-Manipulation Control}.
\newblock In \emph{International Conference on Learning Representations}, 2026.

\bibitem[Jiang et~al.(2025)Jiang, Yuan, Liu, Lu, Cui, Liu, Cheng, Gao, Xu, and Zhao]{jiang2025galaxea}
Tao Jiang, Tianyuan Yuan, Yicheng Liu, Chenhao Lu, Jianning Cui, Xiao Liu, Shuiqi Cheng, Jiyang Gao, Huazhe Xu, and Hang Zhao.
\newblock {Galaxea Open-World Dataset and G0 Dual-System VLA Model}.
\newblock arXiv preprint arXiv:2509.00576, 2025.

\bibitem[Khazatsky et~al.(2024)Khazatsky, Pertsch, Nair, et~al.]{khazatsky2024droid}
Alexander Khazatsky, Karl Pertsch, Suraj Nair, et~al.
\newblock {DROID: A Large-Scale In-The-Wild Robot Manipulation Dataset}.
\newblock arXiv preprint arXiv:2403.12945, 2024.

\bibitem[Kim et~al.(2024)Kim, Pertsch, Karamcheti, Xiao, Balakrishna, Nair, Rafailov, Foster, Lam, Sanketi, Vuong, Kollar, Burchfiel, Tedrake, Sadigh, Levine, Liang, and Finn]{kim2024openvla}
Moo~Jin Kim, Karl Pertsch, Siddharth Karamcheti, Ted Xiao, Ashwin Balakrishna, Suraj Nair, Rafael Rafailov, Ethan Foster, Grace Lam, Pannag Sanketi, Quan Vuong, Thomas Kollar, Benjamin Burchfiel, Russ Tedrake, Dorsa Sadigh, Sergey Levine, Percy Liang, and Chelsea Finn.
\newblock {OpenVLA: An Open-Source Vision-Language-Action Model}.
\newblock In \emph{Conference on Robot Learning}, pages 2679--2713, 2024.

\bibitem[Kostrikov et~al.(2021)Kostrikov, Yarats, and Fergus]{yarats2021drq}
Ilya Kostrikov, Denis Yarats, and Rob Fergus.
\newblock {Image Augmentation Is All You Need: Regularizing Deep Reinforcement Learning from Pixels}.
\newblock In \emph{International Conference on Learning Representations}, 2021.

\bibitem[Krantz et~al.(2020)Krantz, Wijmans, Majumdar, Batra, and Lee]{krantz2020vlnce}
Jacob Krantz, Erik Wijmans, Arjun Majumdar, Dhruv Batra, and Stefan Lee.
\newblock {Beyond the Nav-Graph: Vision-and-Language Navigation in Continuous Environments}.
\newblock In \emph{European Conference on Computer Vision}, pages 104--120, 2020.

\bibitem[Ku et~al.(2020)Ku, Anderson, Patel, Ie, and Baldridge]{ku2020rxr}
Alexander Ku, Peter Anderson, Roma Patel, Eugene Ie, and Jason Baldridge.
\newblock {Room-Across-Room: Multilingual Vision-and-Language Navigation with Dense Spatiotemporal Grounding}.
\newblock In \emph{Conference on Empirical Methods in Natural Language Processing}, pages 4392--4412, 2020.

\bibitem[Laskin et~al.(2020)Laskin, Lee, Stooke, Pinto, Abbeel, and Srinivas]{laskin2020rad}
Michael Laskin, Kimin Lee, Adam Stooke, Lerrel Pinto, Pieter Abbeel, and Aravind Srinivas.
\newblock {Reinforcement Learning with Augmented Data}.
\newblock In \emph{Advances in Neural Information Processing Systems}, 2020.

\bibitem[Lee et~al.(2024)Lee, Wang, Etukuru, Kim, Shafiullah, and Pinto]{lee2024vqbet}
Seungjae Lee, Yibin Wang, Haritheja Etukuru, H.~Jin Kim, Nur Muhammad~Mahi Shafiullah, and Lerrel Pinto.
\newblock {Behavior Generation with Latent Actions}.
\newblock In \emph{Proceedings of the 41st International Conference on Machine Learning}, pages 26991--27008, 2024.

\bibitem[Li et~al.(2026)Li, Chen, Li, Cao, Ren, Tomizuka, and Shah]{li2026tango}
Anqi Li, Yuxin Chen, Zhaobo Li, Zhuo Cao, Junli Ren, Masayoshi Tomizuka, and Dhruv Shah.
\newblock {TANGO: Humanoid Navigation in Cluttered Environments with a Whole-Body Vision-Language-Action Model}.
\newblock arXiv preprint arXiv:2609.09158, 2026.

\bibitem[Liang et~al.(2023)Liang, Huang, Xia, Xu, Hausman, Ichter, Florence, and Zeng]{liang2023codepolicies}
Jacky Liang, Wenlong Huang, Fei Xia, Peng Xu, Karol Hausman, Brian Ichter, Pete Florence, and Andy Zeng.
\newblock {Code as Policies: Language Model Programs for Embodied Control}.
\newblock In \emph{IEEE International Conference on Robotics and Automation}, pages 9493--9500, 2023.

\bibitem[Liao et~al.(2026)Liao, Truong, Huang, Gao, Tevet, Sreenath, and Liu]{liao2025beyondmimic}
Qiayuan Liao, Takara~E. Truong, Xiaoyu Huang, Yuman Gao, Guy Tevet, Koushil Sreenath, and C.~Karen Liu.
\newblock {BeyondMimic: From Motion Tracking to Versatile Humanoid Control via Guided Diffusion}.
\newblock \emph{Science Robotics}, 11\penalty0 (117):\penalty0 eadx8924, 2026.

\bibitem[Lipman et~al.(2023)Lipman, Chen, Ben-Hamu, Nickel, and Le]{lipman2023flow}
Yaron Lipman, Ricky T.~Q. Chen, Heli Ben-Hamu, Maximilian Nickel, and Matt Le.
\newblock {Flow Matching for Generative Modeling}.
\newblock In \emph{International Conference on Learning Representations}, 2023.

\bibitem[Liu et~al.(2024)Liu, Orru, Vakil, Paxton, Shafiullah, and Pinto]{liu2024okrobot}
Peiqi Liu, Yaswanth Orru, Jay Vakil, Chris Paxton, Nur Muhammad~Mahi Shafiullah, and Lerrel Pinto.
\newblock {OK-Robot: What Really Matters in Integrating Open-Knowledge Models for Robotics}.
\newblock arXiv preprint arXiv:2401.12202, 2024.

\bibitem[Luo et~al.(2026)Luo, Yuan, Wang, Li, Casta{\~n}eda, Chen, Cao, Li, Minor, Ben, Park, Sami, Wang, Da, Ding, Hogg, Song, Lim, Jeong, He, Xue, Xiao, Yuen, Kautz, Chang, Iqbal, Fan, and Zhu]{luo2026sonic}
Zhengyi Luo, Ye Yuan, Tingwu Wang, Chenran Li, Fernando Casta{\~n}eda, Sirui Chen, Zi-Ang Cao, Jiefeng Li, David Minor, Qingwei Ben, Jinhyung Park, David Sami, Zi Wang, Xingye Da, Runyu Ding, Cyrus Hogg, Lina Song, Edy Lim, Eugene Jeong, Tairan He, Haoru Xue, Wenli Xiao, Simon Yuen, Jan Kautz, Yan Chang, Umar Iqbal, Linxi~Jim Fan, and Yuke Zhu.
\newblock {SONIC: Supersizing Motion Tracking for Natural Humanoid Whole-Body Control}.
\newblock \emph{Science Robotics}, 11\penalty0 (117):\penalty0 eaed4592, 2026.

\bibitem[Lynch et~al.(2020)Lynch, Khansari, Xiao, Kumar, Tompson, Levine, and Sermanet]{lynch2020latentplans}
Corey Lynch, Mohi Khansari, Ted Xiao, Vikash Kumar, Jonathan Tompson, Sergey Levine, and Pierre Sermanet.
\newblock {Learning Latent Plans from Play}.
\newblock In \emph{Proceedings of the Conference on Robot Learning}, pages 1113--1132, 2020.

\bibitem[Majumdar et~al.(2023)Majumdar, Yadav, Arnaud, Ma, Chen, Silwal, Jain, Berges, Wu, Vakil, Abbeel, Malik, Batra, Lin, Maksymets, Rajeswaran, and Meier]{majumdar2024vc1}
Arjun Majumdar, Karmesh Yadav, Sergio Arnaud, Yecheng~Jason Ma, Claire Chen, Sneha Silwal, Aryan Jain, Vincent-Pierre Berges, Tingfan Wu, Jay Vakil, Pieter Abbeel, Jitendra Malik, Dhruv Batra, Yixin Lin, Oleksandr Maksymets, Aravind Rajeswaran, and Franziska Meier.
\newblock {Where are we in the search for an Artificial Visual Cortex for Embodied Intelligence?}
\newblock In \emph{Advances in Neural Information Processing Systems}, 2023.

\bibitem[McNemar(1947)]{mcnemar1947note}
Quinn McNemar.
\newblock {Note on the sampling error of the difference between correlated proportions or percentages}.
\newblock \emph{Psychometrika}, 12\penalty0 (2):\penalty0 153--157, 1947.

\bibitem[Miki et~al.(2022)Miki, Lee, Hwangbo, Wellhausen, Koltun, and Hutter]{miki2022perceptive}
Takahiro Miki, Joonho Lee, Jemin Hwangbo, Lorenz Wellhausen, Vladlen Koltun, and Marco Hutter.
\newblock {Learning robust perceptive locomotion for quadrupedal robots in the wild}.
\newblock \emph{Science Robotics}, 7\penalty0 (62), 2022.

\bibitem[Nair et~al.(2023)Nair, Rajeswaran, Kumar, Finn, and Gupta]{nair2022r3m}
Suraj Nair, Aravind Rajeswaran, Vikash Kumar, Chelsea Finn, and Abhinav Gupta.
\newblock {R3M: A Universal Visual Representation for Robot Manipulation}.
\newblock In \emph{Proceedings of the 6th Conference on Robot Learning}, pages 892--909, 2023.

\bibitem[{NVIDIA}(2025)]{mittal2025isaaclab}
{NVIDIA}.
\newblock {Isaac Lab: A GPU-Accelerated Simulation Framework for Multi-Modal Robot Learning}.
\newblock arXiv preprint arXiv:2511.04831, 2025.

\bibitem[{NVIDIA} et~al.(2025){NVIDIA}, Bjorck, Casta{\~n}eda, Cherniadev, Da, Ding, Fan, Fang, Fox, Hu, Huang, Jang, Jiang, Kautz, Kundalia, Lao, Li, Lin, Lin, Liu, Llontop, Magne, Mandlekar, Narayan, Nasiriany, Reed, Tan, Wang, Wang, Wang, Wang, Xiang, Xie, Xu, Xu, Ye, Yu, Zhang, Zhang, Zhao, Zheng, and Zhu]{nvidia2025grootn1}
{NVIDIA}, Johan Bjorck, Fernando Casta{\~n}eda, Nikita Cherniadev, Xingye Da, Runyu Ding, Linxi~Jim Fan, Yu Fang, Dieter Fox, Fengyuan Hu, Spencer Huang, Joel Jang, Zhenyu Jiang, Jan Kautz, Kaushil Kundalia, Lawrence Lao, Zhiqi Li, Zongyu Lin, Kevin Lin, Guilin Liu, Edith Llontop, Loic Magne, Ajay Mandlekar, Avnish Narayan, Soroush Nasiriany, Scott Reed, You~Liang Tan, Guanzhi Wang, Zu Wang, Jing Wang, Qi Wang, Jiannan Xiang, Yuqi Xie, Yinzhen Xu, Zhenjia Xu, Seonghyeon Ye, Zhiding Yu, Ao Zhang, Hao Zhang, Yizhou Zhao, Ruijie Zheng, and Yuke Zhu.
\newblock {GR00T N1: An Open Foundation Model for Generalist Humanoid Robots}.
\newblock arXiv preprint arXiv:2503.14734, 2025.

\bibitem[{Open X-Embodiment Collaboration}(2023)]{oneill2024rtx}
{Open X-Embodiment Collaboration}.
\newblock {Open X-Embodiment: Robotic Learning Datasets and RT-X Models}.
\newblock arXiv preprint arXiv:2310.08864, 2023.

\bibitem[Oquab et~al.(2024)Oquab, Darcet, Moutakanni, et~al.]{oquab2024dinov2}
Maxime Oquab, Timoth\'{e}e Darcet, Th\'{e}o Moutakanni, et~al.
\newblock {DINOv2: Learning Robust Visual Features without Supervision}.
\newblock \emph{Transactions on Machine Learning Research}, 2024.

\bibitem[Peng et~al.(2018)Peng, Abbeel, Levine, and van~de Panne]{peng2018deepmimic}
Xue~Bin Peng, Pieter Abbeel, Sergey Levine, and Michiel van~de Panne.
\newblock {DeepMimic: Example-Guided Deep Reinforcement Learning of Physics-Based Character Skills}.
\newblock \emph{ACM Transactions on Graphics}, 37\penalty0 (4):\penalty0 143, 2018.

\bibitem[Peng et~al.(2021)Peng, Ma, Abbeel, Levine, and Kanazawa]{peng2021amp}
Xue~Bin Peng, Ze Ma, Pieter Abbeel, Sergey Levine, and Angjoo Kanazawa.
\newblock {AMP: Adversarial Motion Priors for Stylized Physics-Based Character Control}.
\newblock \emph{ACM Transactions on Graphics}, 40\penalty0 (4):\penalty0 144:1--144:20, 2021.

\bibitem[Peng et~al.(2022)Peng, Guo, Halper, Levine, and Fidler]{peng2022ase}
Xue~Bin Peng, Yunrong Guo, Lina Halper, Sergey Levine, and Sanja Fidler.
\newblock {ASE: Large-Scale Reusable Adversarial Skill Embeddings for Physically Simulated Characters}.
\newblock \emph{ACM Transactions on Graphics}, 41\penalty0 (4):\penalty0 94:1--94:17, 2022.

\bibitem[Radford et~al.(2021)Radford, Kim, Hallacy, Ramesh, Goh, Agarwal, Sastry, Askell, Mishkin, Clark, Krueger, and Sutskever]{radford2021clip}
Alec Radford, Jong~Wook Kim, Chris Hallacy, Aditya Ramesh, Gabriel Goh, Sandhini Agarwal, Girish Sastry, Amanda Askell, Pamela Mishkin, Jack Clark, Gretchen Krueger, and Ilya Sutskever.
\newblock {Learning Transferable Visual Models From Natural Language Supervision}.
\newblock In \emph{International Conference on Machine Learning}, pages 8748--8763, 2021.

\bibitem[Radosavovic et~al.(2023)Radosavovic, Xiao, James, Abbeel, Malik, and Darrell]{radosavovic2023mvp}
Ilija Radosavovic, Tete Xiao, Stephen James, Pieter Abbeel, Jitendra Malik, and Trevor Darrell.
\newblock {Real-World Robot Learning with Masked Visual Pre-training}.
\newblock In \emph{Proceedings of The 6th Conference on Robot Learning}, pages 416--426. PMLR, 2023.

\bibitem[Romero et~al.(2015)Romero, Ballas, Kahou, Chassang, Gatta, and Bengio]{romero2015fitnets}
Adriana Romero, Nicolas Ballas, Samira~Ebrahimi Kahou, Antoine Chassang, Carlo Gatta, and Yoshua Bengio.
\newblock {FitNets: Hints for Thin Deep Nets}.
\newblock In \emph{International Conference on Learning Representations}, 2015.

\bibitem[Rudin et~al.(2022)Rudin, Hoeller, Reist, and Hutter]{rudin2022walk}
Nikita Rudin, David Hoeller, Philipp Reist, and Marco Hutter.
\newblock {Learning to Walk in Minutes Using Massively Parallel Deep Reinforcement Learning}.
\newblock In \emph{Proceedings of the 5th Conference on Robot Learning}, pages 91--100. PMLR, 2022.

\bibitem[Savva et~al.(2019)Savva, Kadian, Maksymets, Zhao, Wijmans, Jain, Straub, Liu, Koltun, Malik, Parikh, and Batra]{savva2019habitat}
Manolis Savva, Abhishek Kadian, Oleksandr Maksymets, Yili Zhao, Erik Wijmans, Bhavana Jain, Julian Straub, Jia Liu, Vladlen Koltun, Jitendra Malik, Devi Parikh, and Dhruv Batra.
\newblock {Habitat: A Platform for Embodied AI Research}.
\newblock In \emph{Proceedings of the IEEE/CVF International Conference on Computer Vision}, pages 9339--9347, 2019.

\bibitem[Shafiullah et~al.(2022)Shafiullah, Cui, Altanzaya, and Pinto]{shafiullah2022bet}
Nur Muhammad~Mahi Shafiullah, Zichen~Jeff Cui, Ariuntuya Altanzaya, and Lerrel Pinto.
\newblock {Behavior Transformers: Cloning $k$ modes with one stone}.
\newblock In \emph{Advances in Neural Information Processing Systems}, 2022.

\bibitem[Shah et~al.(2023{\natexlab{a}})Shah, Sridhar, Bhorkar, Hirose, and Levine]{shah2023gnm}
Dhruv Shah, Ajay Sridhar, Arjun Bhorkar, Noriaki Hirose, and Sergey Levine.
\newblock {GNM: A General Navigation Model to Drive Any Robot}.
\newblock In \emph{IEEE International Conference on Robotics and Automation}, pages 7226--7233, 2023{\natexlab{a}}.

\bibitem[Shah et~al.(2023{\natexlab{b}})Shah, Sridhar, Dashora, Stachowicz, Black, Hirose, and Levine]{shah2023vint}
Dhruv Shah, Ajay Sridhar, Nitish Dashora, Kyle Stachowicz, Kevin Black, Noriaki Hirose, and Sergey Levine.
\newblock {ViNT: A Foundation Model for Visual Navigation}.
\newblock In \emph{Conference on Robot Learning}, pages 711--733, 2023{\natexlab{b}}.

\bibitem[Sridhar et~al.(2024)Sridhar, Shah, Glossop, and Levine]{sridhar2023nomad}
Ajay Sridhar, Dhruv Shah, Catherine Glossop, and Sergey Levine.
\newblock {NoMaD: Goal Masked Diffusion Policies for Navigation and Exploration}.
\newblock In \emph{IEEE International Conference on Robotics and Automation}, pages 63--70, 2024.

\bibitem[Tian et~al.(2025)Tian, Yang, Zeng, Wang, Lin, Dong, and Pang]{tian2024seer}
Yang Tian, Sizhe Yang, Jia Zeng, Ping Wang, Dahua Lin, Hao Dong, and Jiangmiao Pang.
\newblock {Predictive Inverse Dynamics Models are Scalable Learners for Robotic Manipulation}.
\newblock In \emph{International Conference on Learning Representations}, 2025.

\bibitem[Tobin et~al.(2017)Tobin, Fong, Ray, Schneider, Zaremba, and Abbeel]{tobin2017domainrandomization}
Josh Tobin, Rachel Fong, Alex Ray, Jonas Schneider, Wojciech Zaremba, and Pieter Abbeel.
\newblock {Domain Randomization for Transferring Deep Neural Networks from Simulation to the Real World}.
\newblock In \emph{IEEE/RSJ International Conference on Intelligent Robots and Systems}, pages 23--30, 2017.

\bibitem[Walke et~al.(2023)Walke, Black, Zhao, Vuong, Zheng, Hansen-Estruch, He, Myers, Kim, Du, Lee, Fang, Finn, and Levine]{walke2023bridge}
Homer~Rich Walke, Kevin Black, Tony~Z. Zhao, Quan Vuong, Chongyi Zheng, Philippe Hansen-Estruch, Andre~Wang He, Vivek Myers, Moo~Jin Kim, Max Du, Abraham Lee, Kuan Fang, Chelsea Finn, and Sergey Levine.
\newblock {BridgeData V2: A Dataset for Robot Learning at Scale}.
\newblock In \emph{Proceedings of The 7th Conference on Robot Learning}, pages 1723--1736. PMLR, 2023.

\bibitem[Wang et~al.(2023)Wang, Fan, Sun, Zhang, Fei-Fei, Xu, Zhu, and Anandkumar]{wang2023mimicplay}
Chen Wang, Linxi Fan, Jiankai Sun, Ruohan Zhang, Li Fei-Fei, Danfei Xu, Yuke Zhu, and Anima Anandkumar.
\newblock {MimicPlay: Long-Horizon Imitation Learning by Watching Human Play}.
\newblock In \emph{Proceedings of the 7th Conference on Robot Learning}, pages 201--221, 2023.

\bibitem[Wang et~al.(2019)Wang, Huang, Celikyilmaz, Gao, Shen, Wang, Wang, and Zhang]{wang2019rcm}
Xin Wang, Qiuyuan Huang, Asli Celikyilmaz, Jianfeng Gao, Dinghan Shen, Yuan-Fang Wang, William~Yang Wang, and Lei Zhang.
\newblock {Reinforced Cross-Modal Matching and Self-Supervised Imitation Learning for Vision-Language Navigation}.
\newblock In \emph{IEEE/CVF Conference on Computer Vision and Pattern Recognition}, 2019.

\bibitem[Wijmans et~al.(2020)Wijmans, Kadian, Morcos, Lee, Essa, Parikh, Savva, and Batra]{wijmans2020ddppo}
Erik Wijmans, Abhishek Kadian, Ari Morcos, Stefan Lee, Irfan Essa, Devi Parikh, Manolis Savva, and Dhruv Batra.
\newblock {DD-PPO: Learning Near-Perfect PointGoal Navigators from 2.5 Billion Frames}.
\newblock In \emph{International Conference on Learning Representations}, 2020.

\bibitem[Wilcoxon(1945)]{wilcoxon1945individual}
Frank Wilcoxon.
\newblock {Individual Comparisons by Ranking Methods}.
\newblock \emph{Biometrics Bulletin}, 1\penalty0 (6):\penalty0 80--83, 1945.

\bibitem[Wu et~al.(2026)Wu, Zhao, Yang, Nie, Jiang, Sun, Li, Hu, Liu, and Liu]{wu2026eatrstereo}
Songwei Wu, Rui Zhao, Fan Yang, Zhongqiang Nie, Zhiduo Jiang, Wandong Sun, Yuwei Li, Jian Hu, Yang Liu, and Hong Liu.
\newblock {EATR-Stereo: Embodiment-Aware Token Routing of Paired Stereo Evidence for Humanoid Vision-Language-Action Control}.
\newblock arXiv preprint arXiv:2608.17453, 2026.

\bibitem[Ye et~al.(2025)Ye, Jang, Jeon, Joo, Yang, Peng, Mandlekar, Tan, Chao, Lin, Liden, Lee, Gao, Zettlemoyer, Fox, and Seo]{ye2025lapa}
Seonghyeon Ye, Joel Jang, Byeongguk Jeon, Sejune Joo, Jianwei Yang, Baolin Peng, Ajay Mandlekar, Reuben Tan, Yu-Wei Chao, Bill~Yuchen Lin, Lars Liden, Kimin Lee, Jianfeng Gao, Luke Zettlemoyer, Dieter Fox, and Minjoon Seo.
\newblock {Latent Action Pretraining from Videos}.
\newblock In \emph{International Conference on Learning Representations}, 2025.

\bibitem[Zagoruyko and Komodakis(2017)]{zagoruyko2017attention}
Sergey Zagoruyko and Nikos Komodakis.
\newblock {Paying More Attention to Attention: Improving the Performance of Convolutional Neural Networks via Attention Transfer}.
\newblock In \emph{International Conference on Learning Representations}, 2017.

\bibitem[Zhang et~al.(2018)Zhang, Vinyals, Munos, and Bengio]{zhang2018overfitting}
Chiyuan Zhang, Oriol Vinyals, Remi Munos, and Samy Bengio.
\newblock {A Study on Overfitting in Deep Reinforcement Learning}.
\newblock arXiv preprint arXiv:1804.06893, 2018.

\bibitem[Zhao et~al.(2023)Zhao, Kumar, Levine, and Finn]{zhao2023act}
Tony~Z. Zhao, Vikash Kumar, Sergey Levine, and Chelsea Finn.
\newblock {Learning Fine-Grained Bimanual Manipulation with Low-Cost Hardware}.
\newblock In \emph{Robotics: Science and Systems}, 2023.

\bibitem[Zheng et~al.(2026)Zheng, Ma, Fan, Wang, Yang, and Liang]{zheng2026motionwam}
Jia Zheng, Teli Ma, Yudong Fan, Zifan Wang, Shuo Yang, and Junwei Liang.
\newblock {MotionWAM: Towards Foundation World Action Models for Real-Time Humanoid Loco-Manipulation}.
\newblock arXiv preprint arXiv:2606.09215, 2026.

\bibitem[Zitkovich et~al.(2023)Zitkovich, Yu, Xu, et~al.]{zitkovich2023rt2}
Brianna Zitkovich, Tianhe Yu, Sichun Xu, et~al.
\newblock {RT-2: Vision-Language-Action Models Transfer Web Knowledge to Robotic Control}.
\newblock In \emph{Proceedings of The 7th Conference on Robot Learning}, pages 2165--2183. PMLR, 2023.

\end{thebibliography}
\end{document}